\documentclass[lettersize,journal]{IEEEtran}
\usepackage{subfigure}
\usepackage{graphicx}
\usepackage{amsmath}
\usepackage{amssymb}
\usepackage{booktabs}
\usepackage[linesnumbered, ruled]{algorithm2e}
\usepackage{threeparttable}
\SetKwRepeat{Do}{do}{while}
\usepackage[capitalize]{cleveref}
\crefname{section}{Sec.}{Secs.}
\Crefname{section}{Section}{Sections}
\Crefname{table}{Table}{Tables}
\crefname{table}{Tab.}{Tabs.}

\begin{document}
\raggedbottom
\title{Risk Occupancy: A New and Efficient Paradigm through Vehicle-Road-Cloud Collaboration}

\author{Jiaxing Chen, Wei Zhong, Bolin Gao, Yifei Liu, Hengduo Zou, Jiaxi Liu, Yanbo Lu, Jin Huang, Zhihua Zhong 
\thanks{This work was supported by the Key Program of the National Natural Science Foundation of China (52172389) and Hetao Shenzhen-Hong Kong Science and Technology Innovation Cooperation Zone (HZQB-KCZYZ-2021055). The authors are with the School of Vehicle and Mobility, Tsinghua University, Beijing 100084, China, and the State Key Laboratory of Intelligent Green Vehicle and Mobility, Tsinghua University, Beijing 100084, China. Jiaxi Liu is with the Civil and Environmental Engineering Department, University of Wisconsin-Madison, WI, 53715, USA. (e-mail: chenjx23@mails.tsinghua.edu.cn) (Corresponding author: Bolin Gao.)}
}



\maketitle

\begin{abstract}

This paper proposes a novel 4D risk occupancy (RiskOcc) perception paradigm under the Vehicle-Road-Cloud integrated architecture, which unifies object detection and local mapping into a single representation spanning four dimensions: road-surface x and y coordinates, risk, and time. Distinct from conventional grid occupancy and risk field methods, this paradigm adopts an anchor-node-based perception approach with a concise yet reliable risk quantification scheme, enabling flexible and accurate capture of static and dynamic object occupancy states at current and future time steps, with per-occupancy-unit risk quantification. Compared with 3D-Occ, the proposed RiskOcc requires only one layer of data volume and achieves higher perceptual computational efficiency. Visualizations of risk occupancy perception results are presented based on the DAIR-V2X dataset, and a path planning method is developed to verify the usability of the risk occupancy map. Experimental results show that at an initial braking speed of 8 m/s, the model improves safety redundancy by 12.5\% and reduces average deceleration by 5.41\%, enhancing both safety and comfort. 

In summary, this study introduces a novel perception paradigm for intelligent connected vehicle (ICV), providing a compact and interpretable risk representation for cooperative perception and downstream planning in ICV systems.

\end{abstract}


\section{Introduction}
\IEEEPARstart{I}{ntelligent} Connected Vehicles (ICVs) are a key component of the Vehicle-Road-Cloud (VRC) system. Compared with conventional autonomous vehicles, ICVs can utilize roadside and cloud computing capabilities to obtain more comprehensive and accurate environmental information.
Currently, there is no unified standard for perception and risk assessment in intelligent transportation and driving, but basic information, such as object positions and speeds, serves as the main reference. Refining these data into more advanced, cohesive perception forms enables intelligent vehicles to make more intuitive and interpretable decisions, serving as a vital supplement to conventional perception information.

\begin{figure}[t!]

    \subfigure[Traffic Scene]{\includegraphics[height=3cm,width=4.35cm]{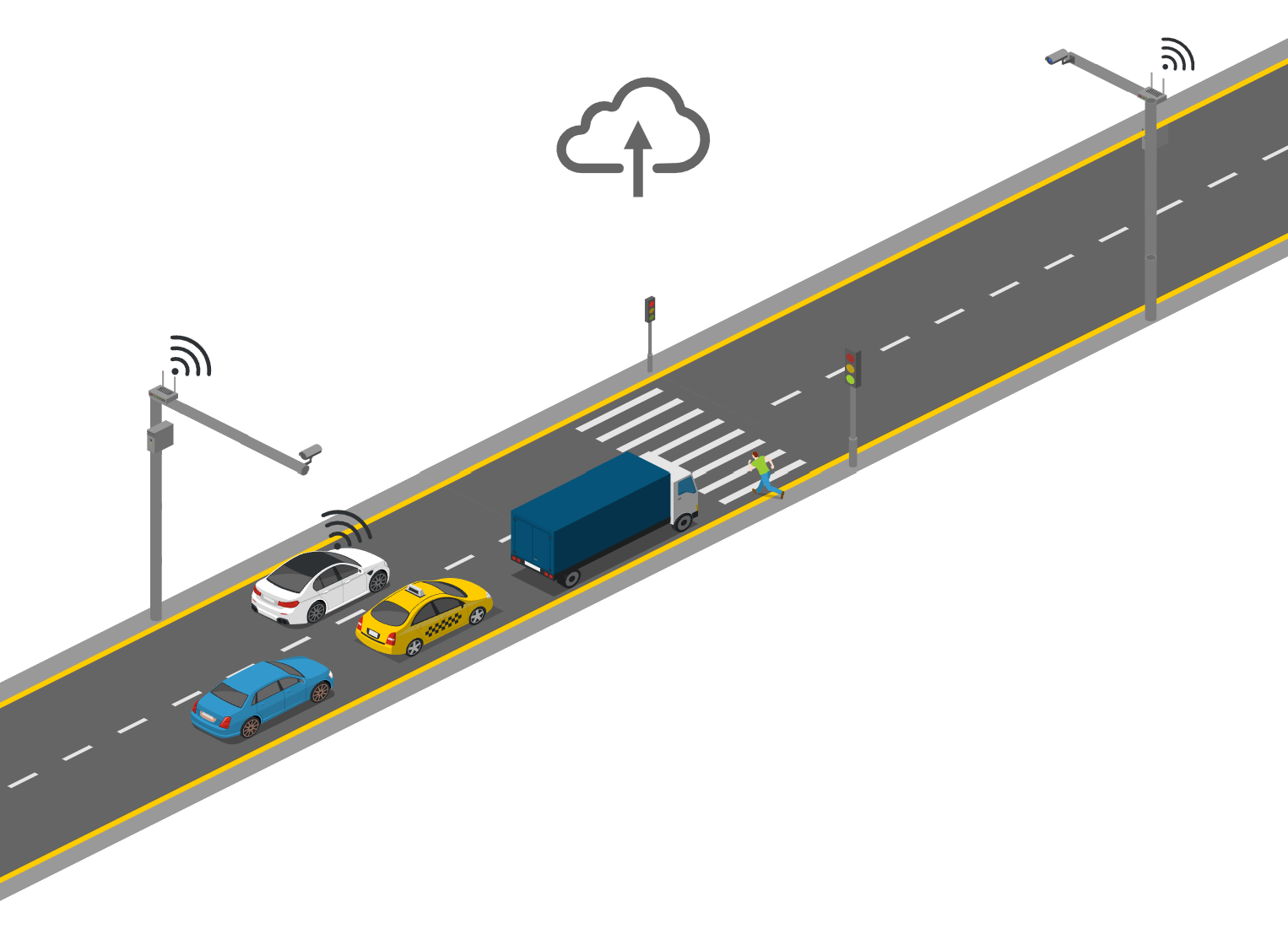}}
    \hfill
    \subfigure[4D Risk Occupancy]{\includegraphics[height=3cm,width=4.35cm]{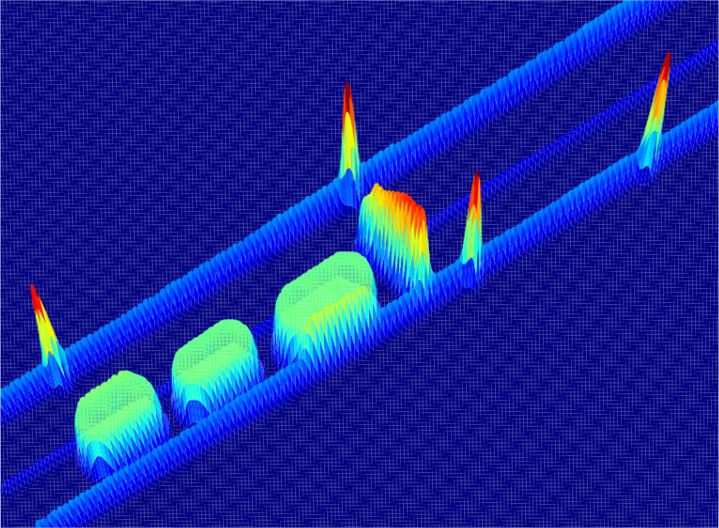}}

    \centering
    \caption{When the green light comes on and a pedestrian runs across the road, the white ICV fails to recognize the runner due to the truck at the front right. Vehicle-Road-Cloud collaboration enables ICVs to access the risk occupancy conditions for approximately the next 3 seconds, perceiving potential risks in advance.}
    \label{first_show}
\end{figure}

Common perception paradigms in the autonomous driving perception field include object detection, local map perception, and methods that combine Bird's Eye View (BEV) object detection \cite{IS-Fusion, EALSS, Bevfusion4D} with local map construction \cite{BEV-LaneDet, StreamMapNet, PVALane}. However, these perception representations are often discrete and inconsistent, reflecting only the current positions of vehicles and lane lines. They are unable to predict future vehicle positions, and their representations are heterogeneous and lack a unified form.

Occupancy grid maps (OGMs) and 3D/4D occupancy methods can divide the road space into units to represent occupancy, but fail to depict lane lines, only reflecting occupancy status. Risk field assessment, though capable of representing lane lines, vehicle information and predicting risks, relies on numerous parameters that are difficult to quantify. Moreover, the unpredictability of driver behavior and the complexity of models introduce computational and storage burdens as well as uncertainty, reducing the algorithm's practicality.

To address this, this paper proposes the 4D Risk Occupancy Perception Paradigm (Fig. \ref{first_show}), which unifies traffic participant and lane line information, refines occupancy risks, and predicts future risk occupancy of each participant. Anchors are placed on a 2D plane via prior road information managed by the edge cloud, with occupancy risks for road anchors constructed using observable data (e.g., speed, heading angle, and object category) and risk values calculated for static and dynamic factors. The paradigm realizes 4D risk occupancy representation by encompassing the $x$- and $y$-spatial dimensions, risk, and time, where the time dimension is reflected through risk values at the unoccupied positions of dynamic objects.

In autonomous driving perception, occlusion poses a threat to driving safety, and existing research on mitigating it has limited effectiveness \cite{occluded}. This paper proposes a solution based on a VRC collaborative architecture that integrates multi-view roadside perception to supplement vehicle-side perception, eliminating occlusions and enhancing reliability. Based on this, paths are planned on the edge cloud accordingly to verify its effectiveness. 

The main contributions are as follows:
\begin{itemize}
    \item We propose a novel 4D risk occupancy perception paradigm that provides a unified representation of traffic participants and lane markings based on observable information. This approach achieves fine-grained differentiation of occupancy states and enables future risk occupancy prediction.
    \item We introduce an anchor-based evaluation methodology that differs from conventional grid/voxel-based approaches. By deploying anchor nodes with prior knowledge, our method forms overlapping perception ranges to create continuous and adaptive perception regions. This design eliminates perception gaps, enables seamless occupancy representation for dynamic objects, and overcomes the limitations of discrete, fixed 3D occupancy representations.
    \item We establish a risk-occupancy-centric cooperative VRC architecture. The system implements fused vehicle-roadside perception to generate cloud-based risk occupancy maps and corresponding path planning solutions.
\end{itemize}

The article begins with an introduction that outlines the research background and significance. The "Related Work" section reviews existing research on perception paradigms. It then explores the risk occupancy perception method and VRC architecture, and proposes a path planning method to validate the effectiveness of the risk occupancy map.

\section{Related Work}

\textbf{Research on 3D-Occ} focuses on lightweight design and cross-domain applications, with algorithms like FlashOcc \cite{flashocc}, TPVFormer \cite{tpvformer}, and VoxelFormer \cite{voxelformer} emerging, while research on 4D-Occ \cite{cam4docc}, \cite{occnet} aims at occupancy prediction and risk assessment in dynamic scenarios, where algorithms such as Occ-LLM \cite{occllm} and OccProphet \cite{occprophet} have made progress in enhancing prediction accuracy and efficiency.

\textbf{An occupancy grid map} divides roads into grid cells to estimate occupancy \cite{ogrids}, including traversable areas and obstacles, built on assumptions of grid independence and static environments during Bayesian filter updates \cite{Zhao_rpbfslam_ogm_2023, Li_mulveh_ogm_merging_2014, toda_2011}. However, these methods fall short in dynamic driving environments with moving objects, which require high OGM update frequencies and spatial resolutions.

\textbf{In driving risk field research}, Wang et al. built a driver-vehicle-road element model \cite{modeling_Wang_safety_field_based_driver_road_vehicle}. As the theory developed, additional elements and methods were added to risk field construction. For example, Han \cite{spatial_temporal_risk_field} used a 3D spatiotemporal risk field for trajectory planning, and other studies incorporated vehicle geometric dimensions, acceleration, and steering angle into the model \cite{sun_exten_app_motion_constraints}, \cite{Li_Gan_Ji_Qu_Ran_2022}. However, most modeling methods depend on parameters that are difficult to observe or quantify, and driver behavior is hard to predict. This results in complex models, high computational costs, and significant uncertainty and noise, severely limiting the paradigm's computational efficiency, practicality, and generality.

\begin{table}[h]
    \centering
    \caption{Comparison of various perception paradigms}
    \label{paradigms}
    \begin{tabular}{c|cc}
    \toprule  \textbf{ } & \textbf{Features of the paradigm}\\ 
    \toprule  \textbf{Det\&Map}	& Representations are discrete, discontinuous, and sparse.   \\ 
    \toprule  \textbf{Occupancy}	& No lane information; occ units mutually independent.    \\ 
    \toprule  \textbf{Risk Field}	& Computationally expensive; unreliable parameter setting.   \\ 
    \toprule  \textbf{Risk-Occ}	& Holistic \& unified representation with inter-anchor sensing.  \\ 
    
    \bottomrule
    \end{tabular}
\end{table}

\textbf{The field of Vehicle-Road-Cloud coordination} is advancing rapidly. Relevant studies \cite{ccs,pta,arch_vrc,vcs,sur_cyber} propose integration solutions, connect vehicles, roadside facilities and cloud resources, ensure communication security, meet multi-scenario needs, adopt cloud-edge-vehicle hierarchical computing, support multi-agent collaboration, and optimize models via big data to enhance system performance. In China, more than 20 pilot programs for the system are currently under testing, with Vehicle-Cloud communication latency of $50$--$500~\mathrm{ms}$ (affected by multiple factors) and vehicles capable of offline autonomous operation. The proposed method requires no extra investment, integrates into mature demonstration zones via existing cloud platforms, and requires improvements to traffic regulations for large-scale application of collaborative algorithms.


Table \ref{paradigms} shows differences among perception approaches.

\section{Modeling Method for 4D Risk Occupancy}

This paper proposes 4D Risk Occupancy as an innovative perception method for the VRC perception architecture. This method integrates traditional perception modalities into a unified, comprehensive, continuous, and intuitive representation of the road environment, providing robust support for downstream planning and decision-making tasks of ICVs. This section describes the modeling techniques of 4D Risk Occupancy, covering key aspects such as risk quantification, anchor-node placement and object-interaction mechanisms, and the assessment process, aiming to provide a complete and concise theoretical framework.

\subsection{Quantification Method for Road Risk}
Road risk quantification is the foundation for constructing 4D Risk Occupancy. The first step in the quantification process is to identify and classify the risk factors that affect the driving safety of ICVs. These risk factors are divided into two major categories: static factors and dynamic factors.

\begin{figure*}[ht] 
 	\centering
        \begin{equation}
            ETA = \frac{Distance_{so}}{Speed} , 
            \label{eq_eta}
        \end{equation}

 	\begin{equation}	    
 		Risk_{dyn} =
 		\begin{cases}
 			 0.0667 \times ETA^3 -0.3 \times ETA^2 + 0.0333 \times ETA + 1,  & \quad ETA~ \leq 3.0,
     
     \\0.5, & \quad ETA ~> 3.0.
 		\end{cases},
 		\label{eq_risk_dyn}
 	\end{equation}
      \begin{equation}	    
     		Risk_{sta} =
     		\begin{cases}
     			 Constant,
                &  Distance_{so} \leq 1.0,
         \\0, & Distance_{so} > 1.0.
     		\end{cases}.
     		\label{eq_risk_sta}
      \end{equation}

 \end{figure*}

Static factors include fixed obstacles such as lane lines, guardrails, curbs, roadblocks, and potholes. Although they are relatively stationary, they can contribute significantly to traffic accidents. This study assigns fixed risk values and weights to each type of static obstacle: curbs and guardrails are given higher weights due to their constraints on vehicle paths, while potholes and lane lines receive lower weights due to their minor impact. Quantifying static factors is crucial for developing a comprehensive risk model for ICV path-planning algorithms, enhancing the safety and reliability of ICV operations.

Dynamic factors, such as human-driven vehicles, non-motorized vehicles, and pedestrians, pose a greater threat to ICVs' safety due to their changing characteristics over time and space. This study quantifies the risks of dynamic factors using critical information such as location, speed, and heading angle, combined with the Estimated Time of Arrival (ETA) metric. The study prioritizes personal safety and minimizes property damage by assigning the highest risk weights to pedestrians and non-motorized vehicles. Larger vehicles receive higher weights than smaller ones, as collisions with larger vehicles cause more harm to ICVs' passengers. Dynamic factors are given higher weights than static factors due to their uncertainty, and this distribution helps ICVs prioritize risk scenarios and select lower-risk evasive strategies.

A safety time threshold of $2.5$ seconds is set to account for the reaction and braking time required by most drivers. Under urban road conditions with a speed limit of $70~\mathrm{km/h}$, autonomous braking typically takes less than $1.5$ seconds. Once the ETA exceeds $1.5$ seconds, the risk value significantly decreases, and the change in risk value diminishes as ETA increases. Experiments show that assigning a risk value of 0.5 to stationary vehicles reflects their higher risk compared to static factors, consistent with risk quantification principles.

\begin{figure}[h]
    \centering
    \includegraphics[width=\linewidth]{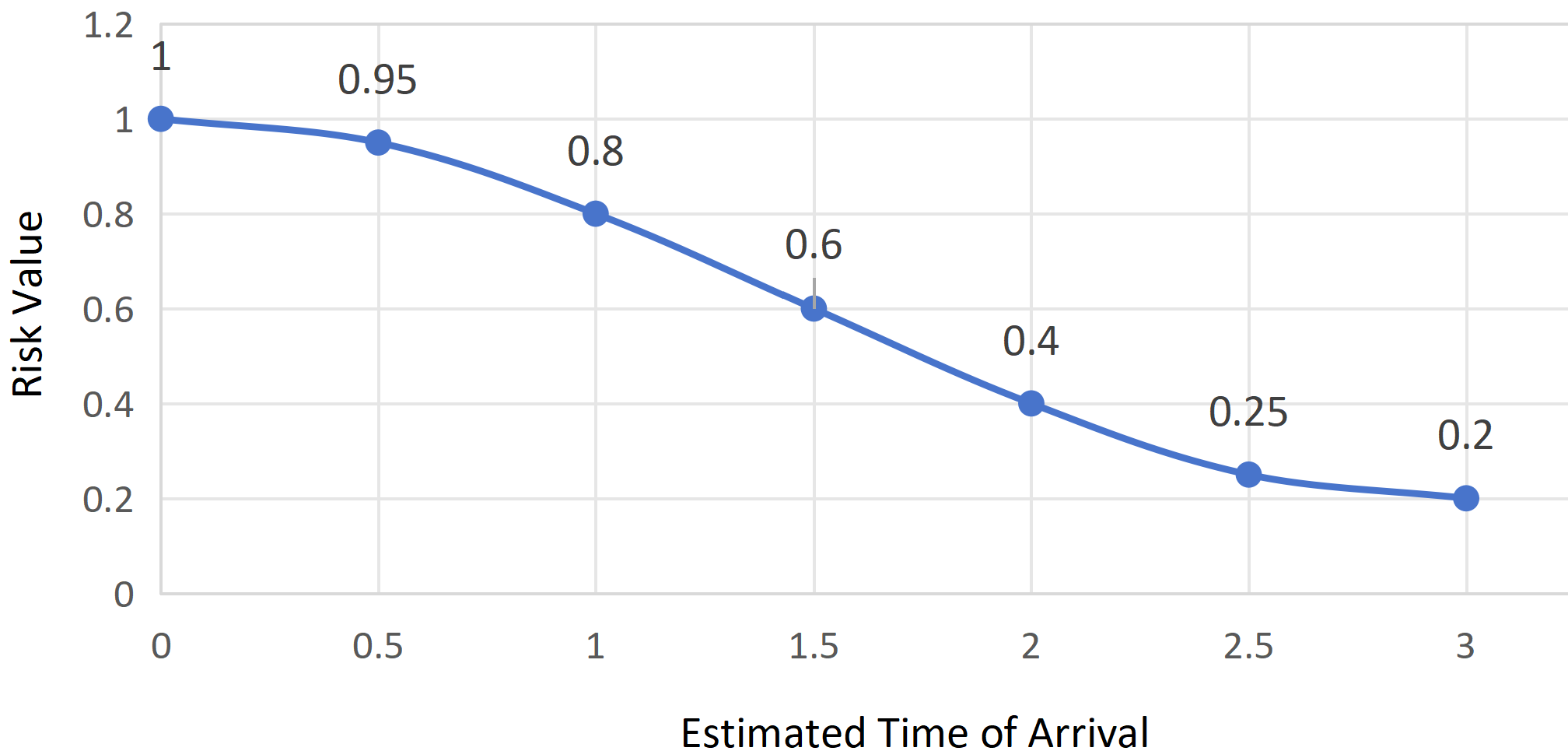}
    \caption{The curvilinear relationship between risk value and ETA.}
    \label{risk-eta}
\end{figure}

Eqs. (\ref{eq_eta}) and (\ref{eq_risk_dyn}) present the specific formulas for risk quantification. Fig. \ref{risk-eta} illustrates the nonlinear negative correlation between ETA and risk values through multiple data points and nonlinear regression. The risk values are normalized to a range of $0$--$1$, consistent with our risk-quantification principles.


\subsection{Anchor Nodes and Geometric Object Interaction Method}

To accurately assess road risks, dynamic and static factors are first converted into corresponding geometric objects, and a method based on the interaction between anchor nodes and geometric objects is designed. The placement strategy of anchor nodes and the interaction mechanism are crucial to ensuring the accuracy of risk assessment.

\begin{figure*}[t]
    \centering    \includegraphics[width=\linewidth]{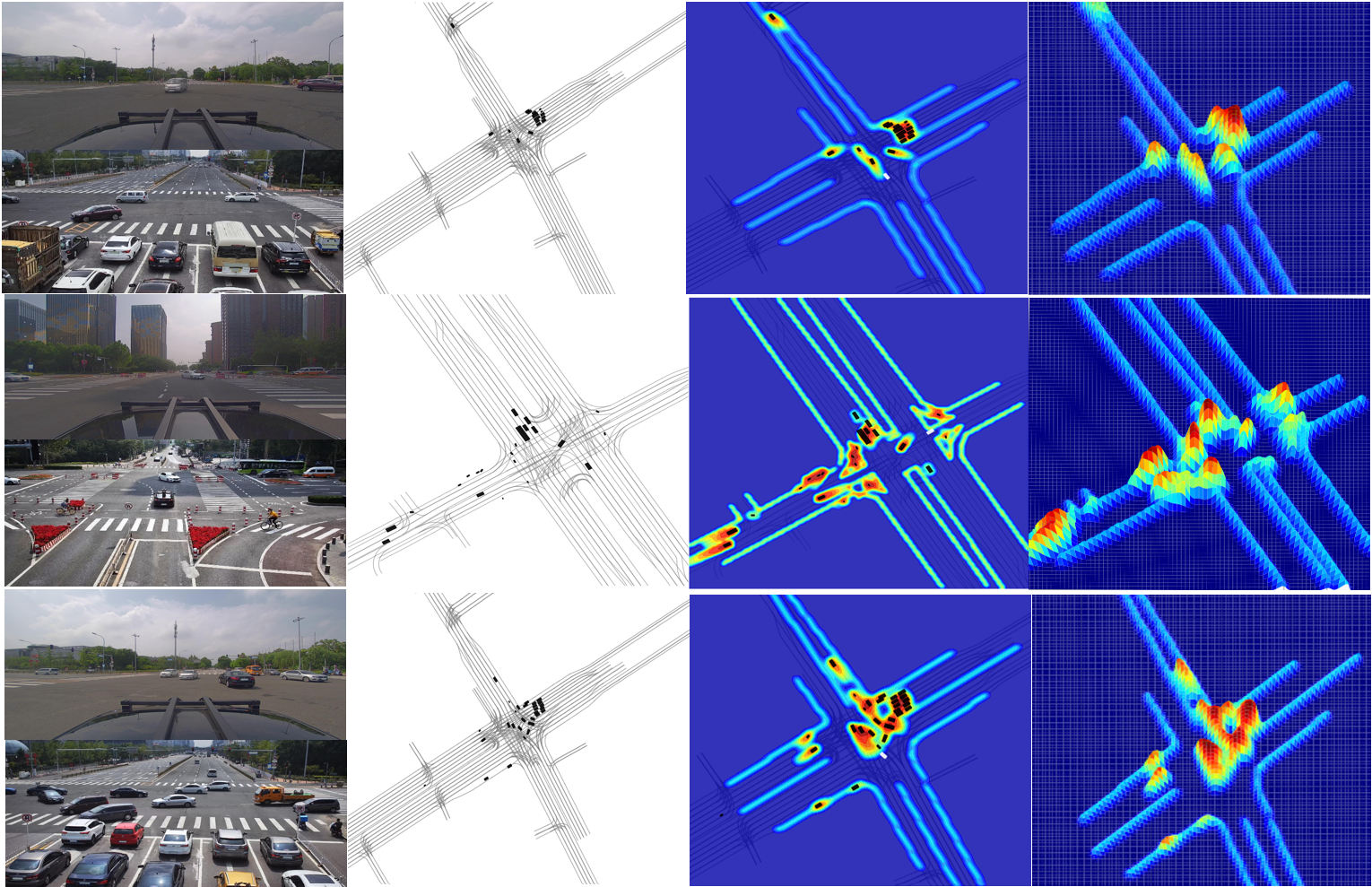}
    \caption{The figure presents the risk occupancy results of three scenarios at two intersections in DAIR-V2X-Seq, a cooperative vehicle-infrastructure dataset.}
	\label{4D Risk Occupancy show}
\end{figure*}


\begin{figure}[h]
    \centering    \includegraphics[width=\linewidth]{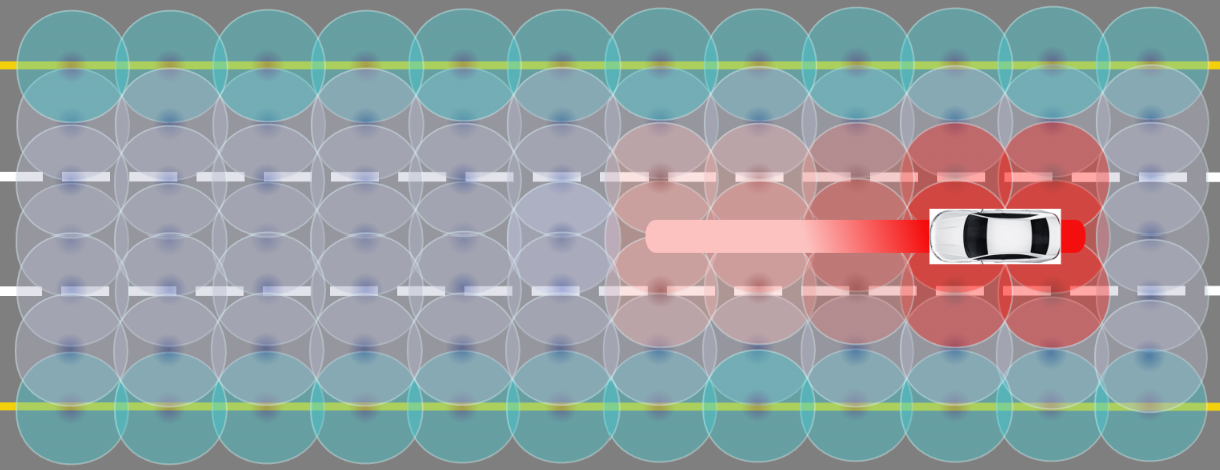}
    \caption{Demonstration of the road anchor node layout. The anchor nodes on the road are arranged along the direction of the road with an adjustable resolution. This figure intuitively demonstrates the detection range of a moving object and its influence on nearby anchor nodes.  }
    \label{sample_assignment}
\end{figure}

The placement of anchor nodes should consider the width, direction, and resolution of urban roads. This study uses a uniform distribution strategy to place anchor nodes at a grid resolution of $1.9~\mathrm{m}$ (Fig. \ref{sample_assignment}). Each anchor node assesses dynamic factors within a 2-meter range and static factors within a 1-meter range, enabling efficient risk evaluation on the road. Traffic risk factors are represented as geometric objects: single points for object locations, multiple points for large-area obstacles, line segments for future positions, and vectors for the direction and layout of lane lines, curbs, and guardrails.

\begin{algorithm}[h]
  \label{algorithm1}
  \caption{Risk Calculation}
  \KwData{Collected from the Vehicle-Road-Cloud system}
  \KwResult{The calculated risk value}
  initialization\;  
    \For{each $dyn$ \textbf{in} $dyns\_info$}{
      \If{$distance \leq 2.0$}{
         $ETA \leftarrow dis\_point_{dyn} / (speed_{dyn} + 0.01)$\;
        \eIf{$ETA \leq 3.0$}{
          $degree\_risk_{dyn} \leftarrow 0.0667 \times ETA^3 - 0.3 \times ETA^2 + 0.0333 \times ETA + 1$\;
        }{
          $degree\_risk_{dyn} \leftarrow 0.5$\;
        }
        $degree\_risk \leftarrow degree\_risk + degree\_risk_{dyn}$\;
      }
    }
    \For{each $stat$ \textbf{in} $stats\_info$}{
      \If{$distance < 1.0$}{
        $degree\_risk \leftarrow degree\_risk + degree\_risk_{stat}$\;
      }
    }
    \If{$degree\_risk>1.0$}{$degree\_risk\leftarrow 1.0$}
    \Return $degree\_risk$\;
\end{algorithm}


During the interaction process, anchor nodes employ differentiated sensing radius strategies for static and dynamic objects: A smaller sensing radius is adopted for static objects to minimize localization errors and ensure more accurate risk occupancy positioning. For dynamic objects, the sensing radius is appropriately enlarged to capture variations in dynamic objects, enabling more responsive prediction of their risk occupancy status while eliminating ``voids'' in occupancy units. This adaptive sensing radius adjustment allows anchor nodes to implement corresponding interaction strategies based on object dynamics, achieving both precise identification of static objects and sensitive detection of dynamic object behaviors.

For dynamic factors, \cite{spatial_temporal_risk_field} noted that the spatial dimension alone cannot fully capture dynamic traffic changes; the time dimension is needed for 3D space-time risk representation. Future risk prediction for dynamic objects is crucial for ICVs. Static factors are detected when vectors/points enter the anchor node's static perception range and are assigned a fixed risk value ($0$--$1$) and a weight based on the type of static factor, as in Eq. (\ref{eq_risk_sta}).

Unlike \cite{spatial_temporal_risk_field}, this study expresses 3D spatiotemporal information on a 2D plane to save resources. The 4D Risk Occupancy predicts the future risk over the next 3 seconds for each dynamic object. Conservatively, it determines the future position by assuming constant speed and direction for $3~\mathrm{s}$, connecting it to the current position to form a line segment. When geometric objects enter the anchor node's dynamic perception range, ETA is calculated as in Eq. (\ref{eq_eta}), where $Distance_{so}$ is the anchor-object distance, and $Speed$ is the object's speed. Polynomial fitting yields the Risk-ETA curve (Fig. \ref{risk-eta} and Eq. (\ref{eq_risk_dyn})) for potential collision risk calculation.

\subsection{The Principle of 4D Risk Occupancy}

The principle of 4D Risk Occupancy involves accumulating the risk values corresponding to multiple dynamic and static factors within the perception range of the anchor nodes to obtain the final road risk value. This process involves traversing the collection of anchor nodes and calculating the risk values.

During the assessment process, this paper first defines the perception range of anchor nodes and adopts corresponding interaction and calculation strategies based on whether the objects are static or dynamic within that range. For dynamic objects, a probe is assigned to each dynamic object to identify relevant anchor nodes for calculation (Fig. \ref{sample_assignment}). The ETA value is then computed, and potential collision risks are assessed based on it. For static objects, fixed risk values and weights are directly assigned according to their type. The pseudocode for calculating 4D Risk Occupancy is presented in Algorithm \ref{algorithm1}.
After calculating the total risk value for each anchor node, this study aggregates and visualizes the risk values of all nodes within the road area, yielding a unified risk occupancy representation from a BEV perspective. This enables an intuitive assessment of current and future driving risks in the environment.

Fig. \ref{4D Risk Occupancy show} illustrates the traffic situation in several scenes from the DAIR-V2X dataset \cite{DAIR_Yu_Luo_et_aL_2022}, which is a large-scale V2X dataset that contains over 15,000 frames of sequential perception data and approximately 210,000 trajectory-prediction scenarios. It covers vehicle-side, roadside, and cooperative V2X perspectives, and includes natural scene data spanning 672 hours. 

\section{A Path Planning Approach Based on 4D Risk Occupancy}

To verify the applicability of the proposed risk occupancy representation, this work develops an anchor-node-constrained rule-based path planning method for validation.

Two mainstream planning paradigms for risk-aware navigation are analyzed. The heuristic A* algorithm \cite{ASTAR_Raphael_1968} generates optimal paths via node cost evaluation, and was enhanced with elliptical risk constraints for directional safety by Tian et al. \cite{Tian_Pei_Zhang_2020}, yet it suffers from cumbersome cost tuning, strong map dependence, and high computational load in dynamic scenarios.
Sampling-based RRT \cite{RRT_LaValle_1998} and its RRT* variant \cite{RRTSTAR_Karaman_Frazzoli_2010} adapt well to high-dimensional complex environments \cite{3D_Traj_plan_RRTST_Pharpatara_2017} through random tree expansion, and have been extended to spatio-temporal risk field planning \cite{spatial_temporal_risk_field}. 

However, their stochastic sampling leads to low efficiency in narrow spaces and produces paths with redundant nodes and unnecessary maneuvers.
Given its strength in fixed start-end tasks and native compatibility with risk values, we adopt an A*-derived planner and narrow its search scope via traffic rules and the anchor node matrix to improve efficiency.

\subsection{Collision-free Dynamic Node Sets}

The path planning algorithm is driven by 4D Risk Occupancy within a VRC collaborative framework. On the edge cloud platform, future driving intentions can be inferred from commands issued by the ICV. Subsequently, path planning is conducted by leveraging the ICV's current location, prior knowledge of lane information, and the assessment outcomes of the 4D Risk Occupancy model.

\begin{figure}[t]
	
	\begin{minipage}{0.32\linewidth}
		\vspace{3pt}
		\centerline{\includegraphics[width=\textwidth]{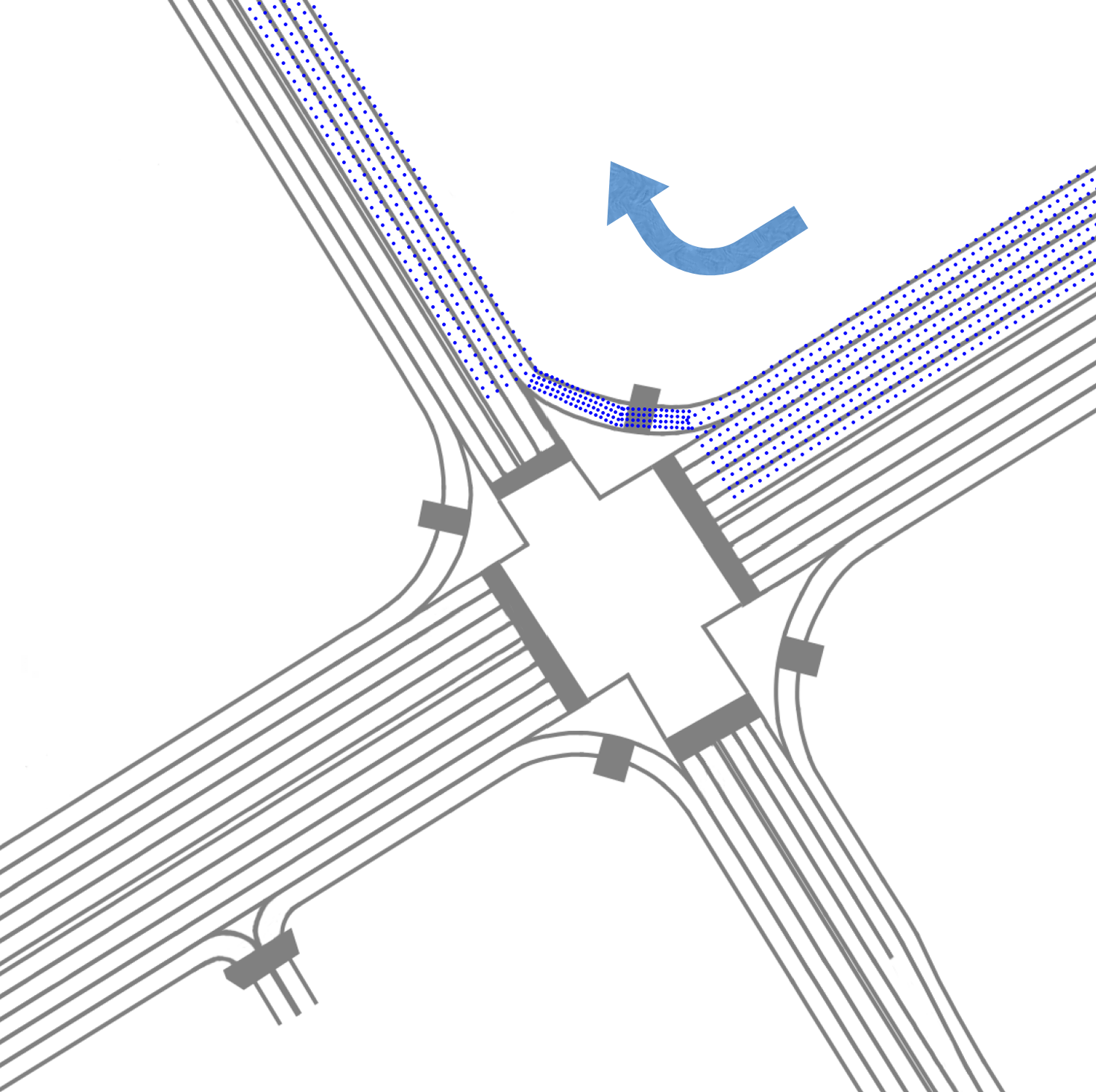}}

	\end{minipage}
	\begin{minipage}{0.32\linewidth}
		\vspace{3pt}
		\centerline{\includegraphics[width=\textwidth]{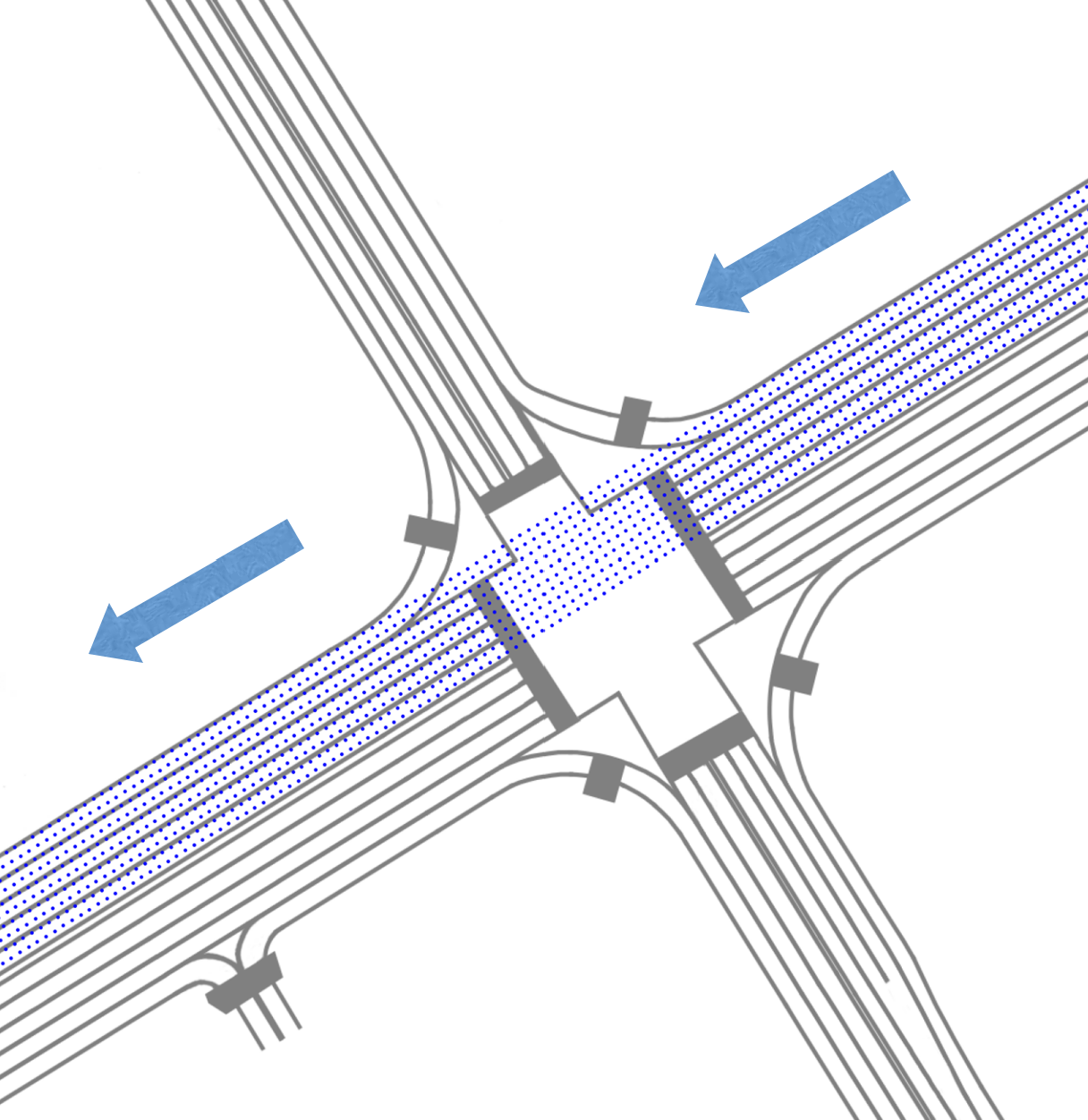}}
	 
	\end{minipage}
	\begin{minipage}{0.32\linewidth}
		\vspace{3pt}
		\centerline{\includegraphics[width=\textwidth]{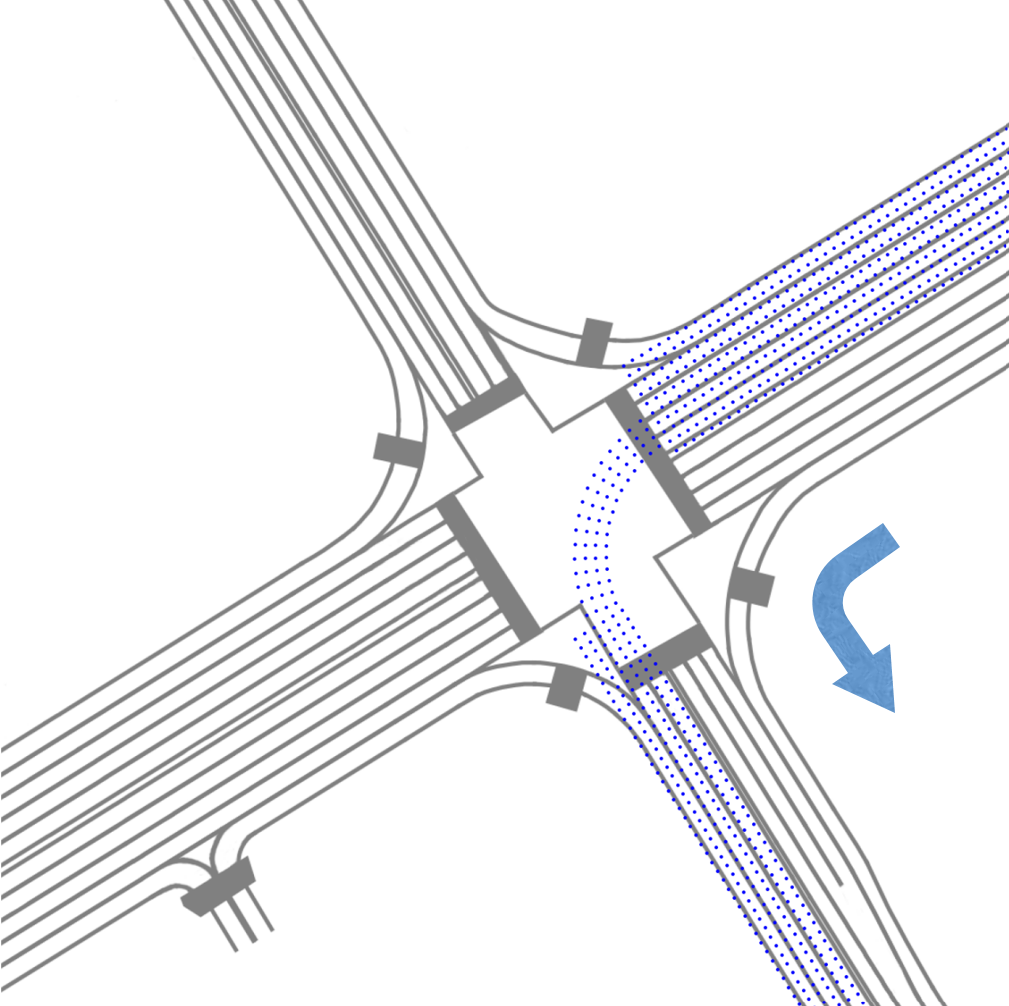}}
	 
	\end{minipage}
 
	\caption{The three figures show the cloud platform's preset potential node sets for an ICV to navigate intersections, corresponding to three driving scenarios: right turns, straight movements, and left turns.}
	\label{Specific node set for 3 kinds of behavior}
\end{figure}

The algorithm starts by preemptively retrieving a set of road points corresponding to the driving behavior instructions, based on the ICV's current location, operational directives, and the road segment in which it is situated. The selection scope of this node set is predicated on the areas that can potentially be reached under the permissible conditions stipulated by traffic regulations. Fig. \ref{Specific node set for 3 kinds of behavior} illustrates the node sets for right turns, straight movements, and left turns that are stored on the edge cloud. When the ICV enters the intersection area, the edge cloud can extract the required preset node set based on the behavior instruction sent by the ICV. Subsequently, anchor nodes that fall below the preset risk threshold are assembled into a collision-free dynamic node set, updated with each frame to support dynamic planning.

\begin{figure*}[t!]
    \centering
    \includegraphics[width=\linewidth]{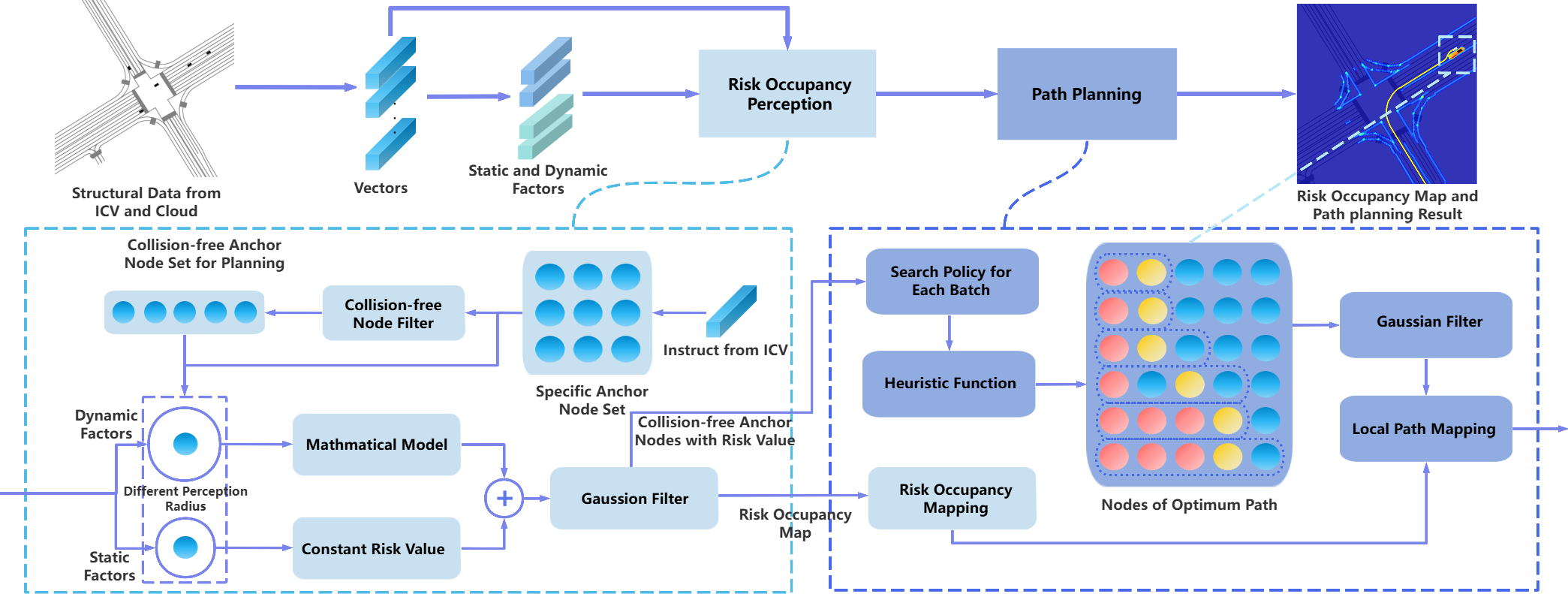}
    \caption{The structural principle of the 4D Risk Occupancy method and a demonstration of a path planning method using risk occupancy.}
    \label{Model}
\end{figure*}

\subsection{Path Planning Strategy}

For computational efficiency, this study constructs a Frenet coordinate system based on the road centerline for the ICV to search for and select the next batch of path nodes. At the same time, a heuristic function is designed in the Cartesian coordinate system to guide the path search. Unlike the conventional A* algorithm, the proposed method uses the Euclidean distance from the nodes to the endpoint in the Cartesian coordinate system and the Euclidean distance from the starting point to the endpoint. In addition, the road risk value is introduced into the cost calculation of the heuristic function. The node with the lowest heuristic function value in each row will be given the highest priority and added to the closed set. The heuristic function is as follows:

\begin{equation}
\begin{split}
    Score = & weight_{risk} \times risk + weight_{dis} \times \\
         &\frac{\sqrt{(x_{node} - x_{dest})^2 + (y_{node} - y_{dest})^2}}{\sqrt{(x_{ICVs} - x_{dest})^2 + (y_{ICVs} - y_{dest})^2}},
\end{split}
\end{equation}
where $(x_{node}, y_{node})$, $(x_{ICVs},y_{ICVs})$ and $(x_{dest},y_{dest})$ denote the coordinates of a node, the ICV, and the destination point, respectively. The weights for the risk and distance costs can be set according to empirical performance.

\textbf{Global Optimization Strategy}: This approach determines a predefined endpoint based on anticipated future driving behaviors and acquires a set of collision-free dynamic nodes for the current state. Reachable nodes are identified according to the position and orientation of the ICV. The costs of all neighboring nodes are computed, and the path with the lowest total cost is selected. In contrast, the local optimization strategy selects only the node with the minimum cost from each batch of neighboring nodes, eliminating the need to compute costs for all reachable nodes, thereby reducing computational load. However, this strategy may lead to suboptimal solutions (Fig. \ref{Model}).

\textbf{Nodes and Paths}: Anchor nodes serve as vertices, and the paths between two anchor nodes are considered edges. Starting from the ICV's position, the next batch of neighboring nodes is determined according to predefined rules, and this process is repeated until the endpoint is reached or no feasible nodes remain. To this end, rules are designed to determine the next batch of neighboring nodes. The node set is numbered based on prior knowledge of road positions. Movement rules are established by integrating the ICV's future behavior and traffic regulations, which narrow the search scope and reduce computational demands. For instance, when the ICV needs to make a left turn, only the nodes corresponding to straight-ahead and left-turn directions are selected, guiding the vehicle to shift left in advance (Fig. \ref{Model}).

\textbf{Path Smoothing}: The initial path generated by the 4D Risk Occupancy Model consists of short straight-line segments, which lack smoothness. A Gaussian path-smoothing technique is employed to address this issue. Gaussian filtering is applied to the original path nodes to generate smoothed nodes through weighted averaging. These smoothed nodes are then refined using interpolation techniques (e.g., spline or polynomial interpolation) to produce a smooth and continuous path, thereby enhancing vehicle driving comfort and efficiency.

\section{Design of Vehicle-Road-Cloud Collaboration Architecture for Risk Occupancy}

This section presents a VRC collaborative architecture for risk occupancy perception. It leverages the resources of vehicles, roads, and the cloud to provide ICVs with a risk occupancy map at intersections, supporting their downstream tasks.


In the VRC collaboration architecture, the risk occupancy algorithm on the cloud platform is engineered to receive diverse data from ICVs, roadside infrastructure, and cloud services, focusing on serving ICVs by providing real-time risk occupancy assessment and path planning for their respective road regions. The system initially constructs a structured dataset including static elements using prior information and integrates dynamic data collected by roadside perception devices, such as the type, location, velocity, and heading angle of traffic participants. These data are transmitted to the edge cloud server, where they undergo a coordinate transformation from the geodetic coordinate system to the Cartesian coordinate system, and ICVs' uploaded locations and instructions are identified and used. This algorithm selectively displays the risk occupancy of local road areas to enhance computational efficiency, completes path planning based on road-risk occupancy, and transmits the risk occupancy information back to ICVs to support their decision-making.

\begin{figure}[t]
    \centering
    \includegraphics[width=\linewidth]{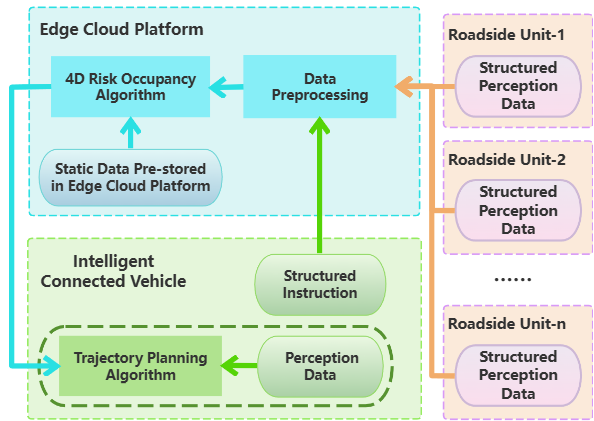}
    \caption{Flowchart of the 4D Risk Occupancy algorithm under the Vehicle-Road-Cloud collaboration architecture.}
    \label{VIC}
\end{figure}

As shown in Fig. \ref{VIC}, the dashed box on the vehicle side is for illustrative purposes only and is not part of the algorithm development in this paper. The output of our risk occupancy perception algorithm can serve as input for subsequent vehicle-side algorithms, expanding application scenarios. While the development and implementation of these subsequent algorithms are beyond the scope of this study, they demonstrate the multifunctionality and scalability of our algorithm's output, which will be explored in future research.

The right side of Fig. \ref{VIC} shows multiple roadside perception units, highlighting the cloud-based algorithm's advantage of obtaining perception information from multiple roadside units to achieve beyond-line-of-sight and global information perception, which cannot be achieved by a single vehicle-side unit.

Within the VRC collaboration framework, the perception algorithm integrates cloud-based prior information with dynamic data from vehicles and roadside sources, reducing dependence on computational and storage resources and enhancing data processing efficiency. Multi-node roadside perception devices provide the algorithm with beyond-line-of-sight and panoramic road information, improving its understanding and prediction of traffic conditions. The edge cloud platform's high-performance computing and low-latency support enable rapid data processing for real-time decision-making and response, which is critical for the adaptability and safety of ICVs.

In summary, the perception algorithm under the VRC collaboration framework constructs an efficient, flexible, and responsive intelligent transportation system by integrating information, providing comprehensive monitoring, and leveraging high-performance computing. This supports decision-making for ICVs and lays a foundation for the future development of intelligent transportation systems.

\begin{figure*}[ht!]
    \centering
    
        \raisebox{0\height}{\rotatebox{90}{\fontsize{9pt}{0pt}\selectfont Right}}\hspace{0pt}
	\begin{minipage}{0.24\linewidth}
		\vspace{3pt}
		\centerline{\includegraphics[width=\textwidth]{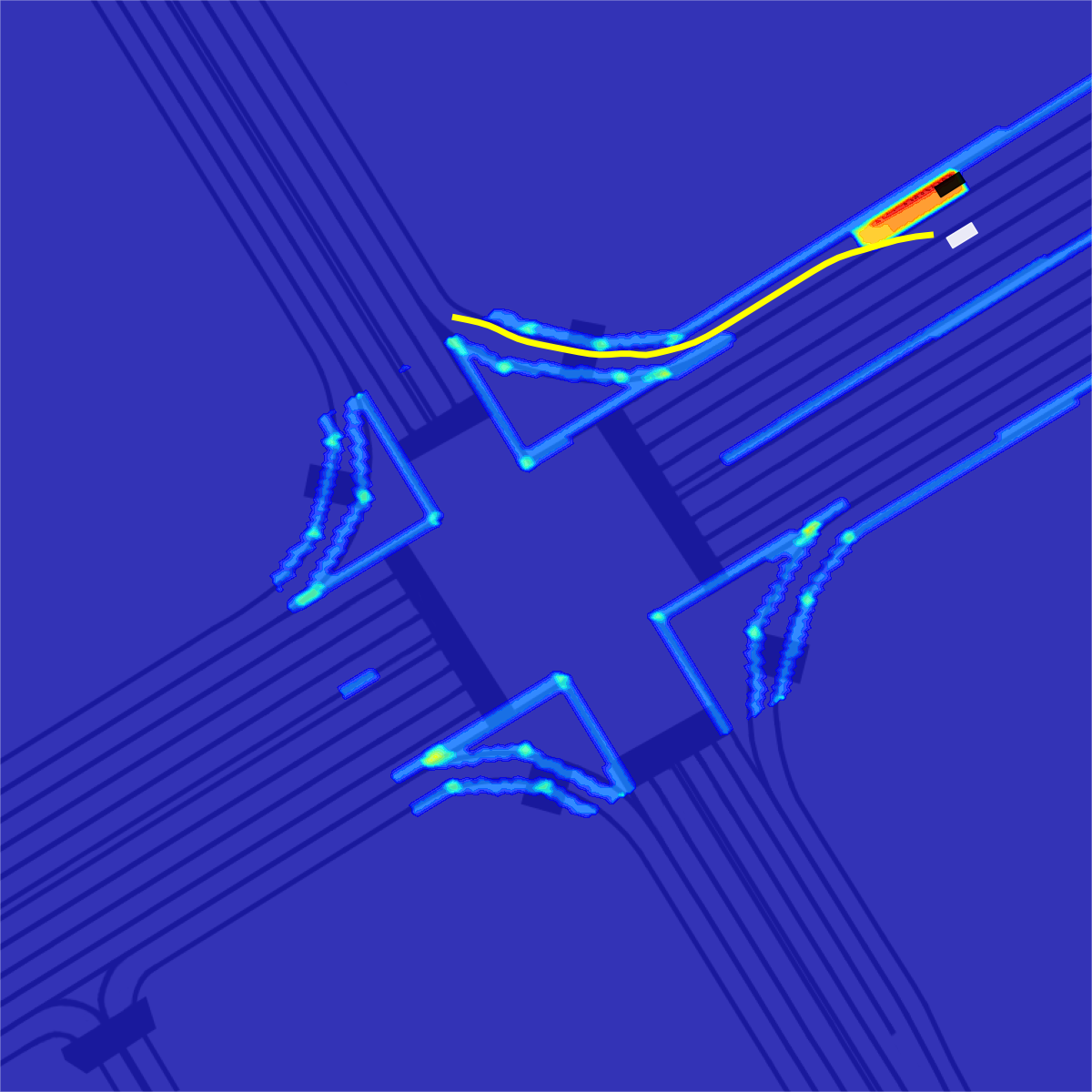}}

	\end{minipage}
	\begin{minipage}{0.24\linewidth}
		\vspace{3pt}
		\centerline{\includegraphics[width=\textwidth]{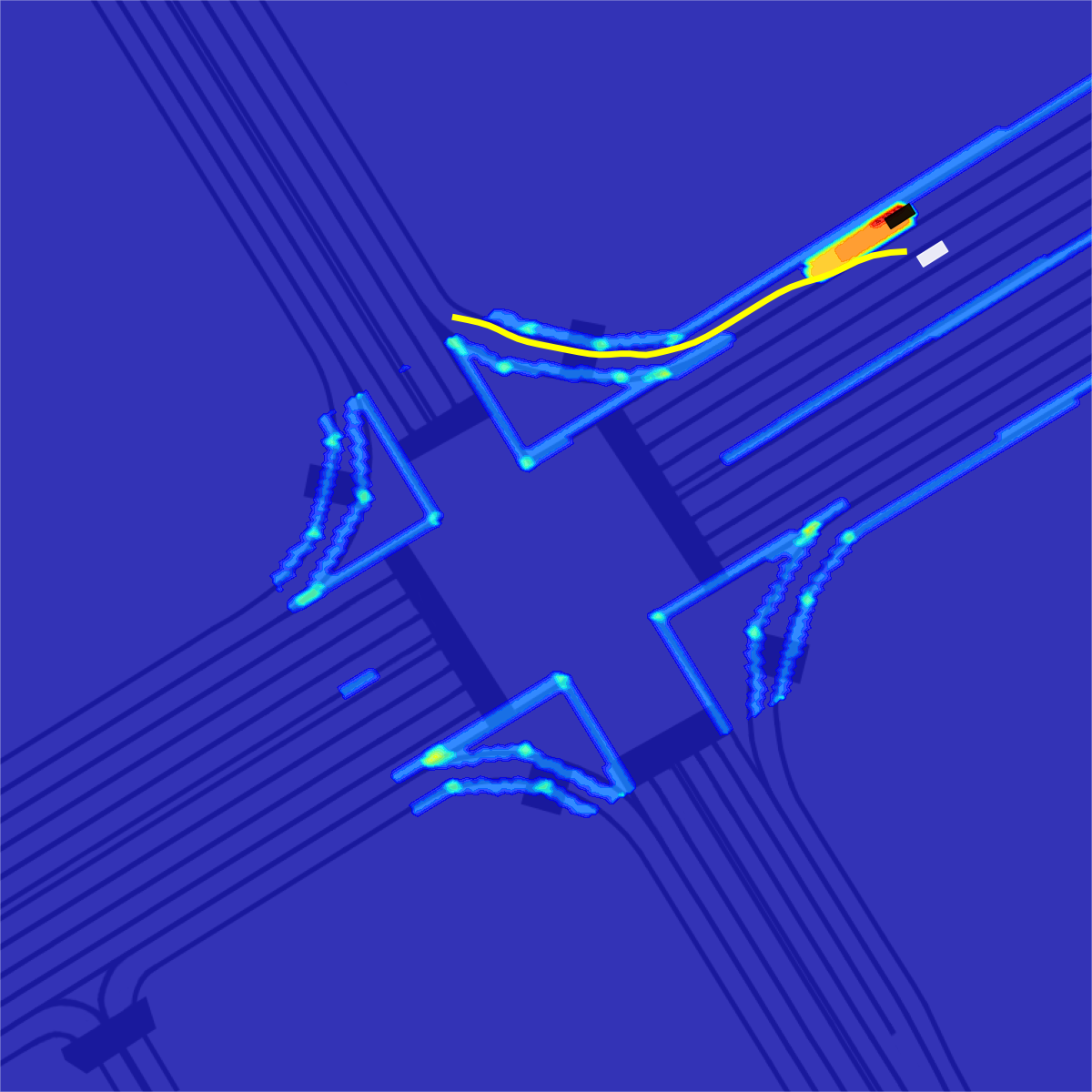}}
	 
	\end{minipage}
	\begin{minipage}{0.24\linewidth}
		\vspace{3pt}
		\centerline{\includegraphics[width=\textwidth]{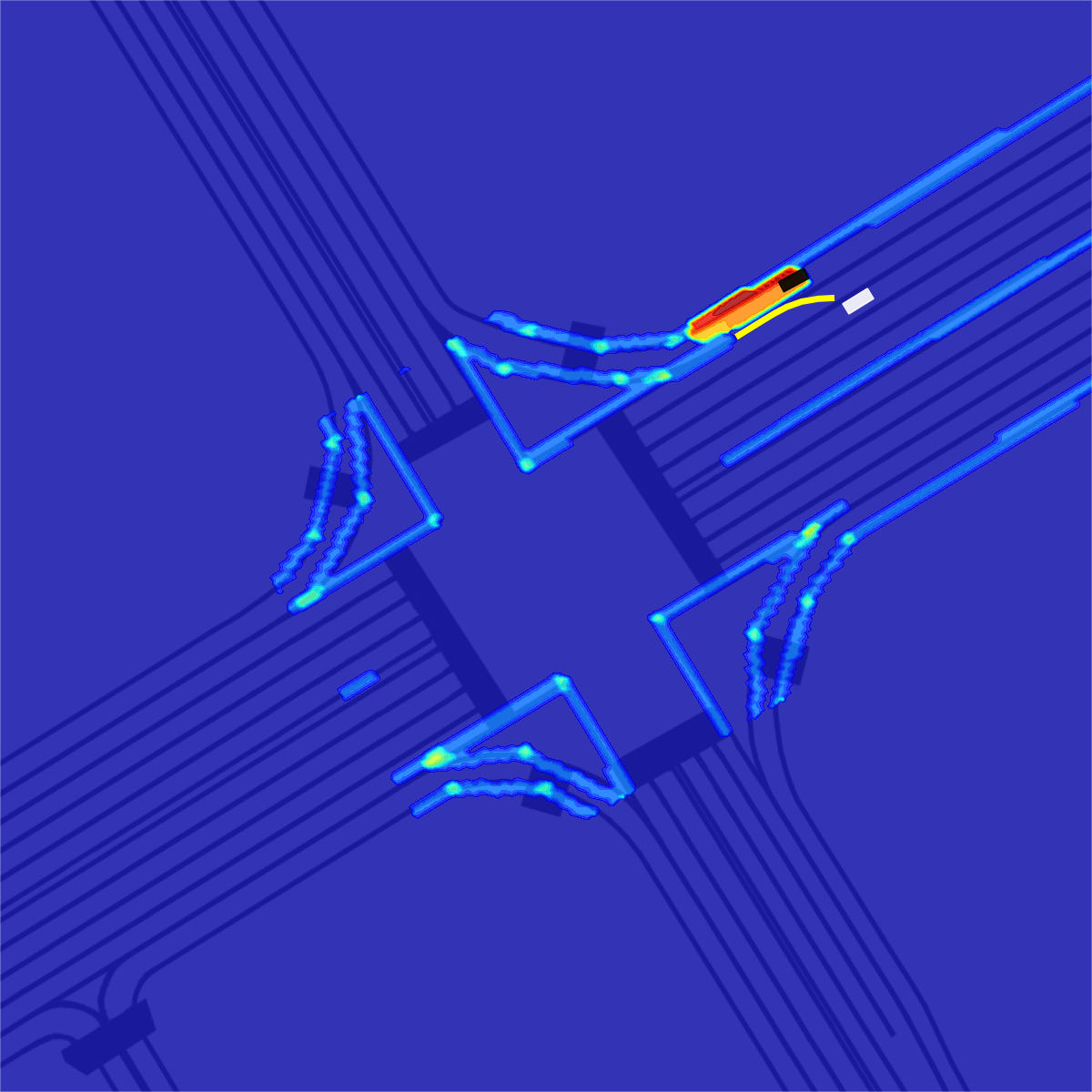}}
	 
	\end{minipage}
    \begin{minipage}{0.24\linewidth}
		\vspace{3pt}
		\centerline{\includegraphics[width=\textwidth]{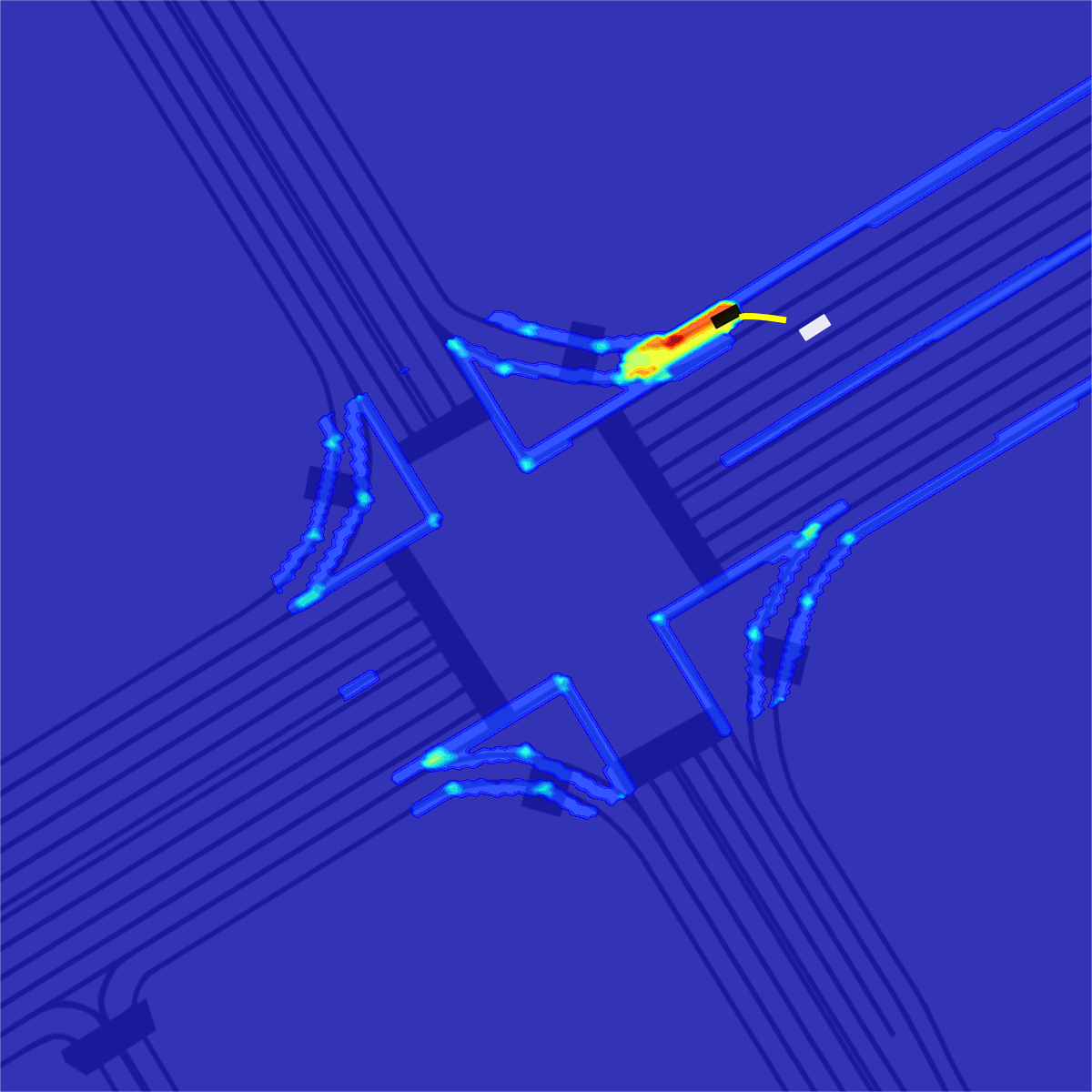}}
	\end{minipage}
 
    \raisebox{0\height}{\rotatebox{90}{\fontsize{9pt}{0pt}\selectfont Right}}\hspace{0pt}
    \begin{minipage}{0.24\linewidth}
		\vspace{3pt}
		\centerline{\includegraphics[width=\textwidth]{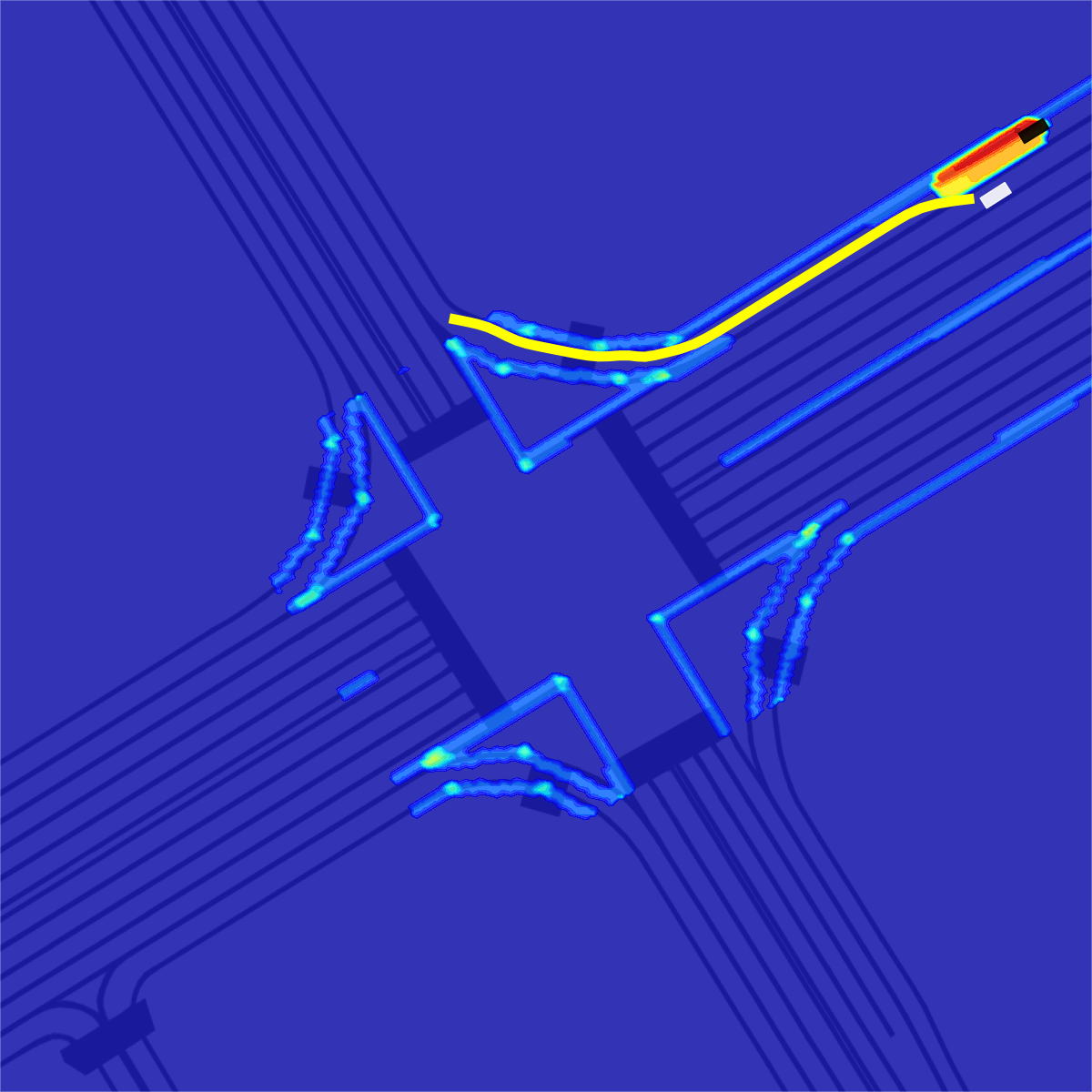}}

	\end{minipage}
	\begin{minipage}{0.24\linewidth}
		\vspace{3pt}
		\centerline{\includegraphics[width=\textwidth]{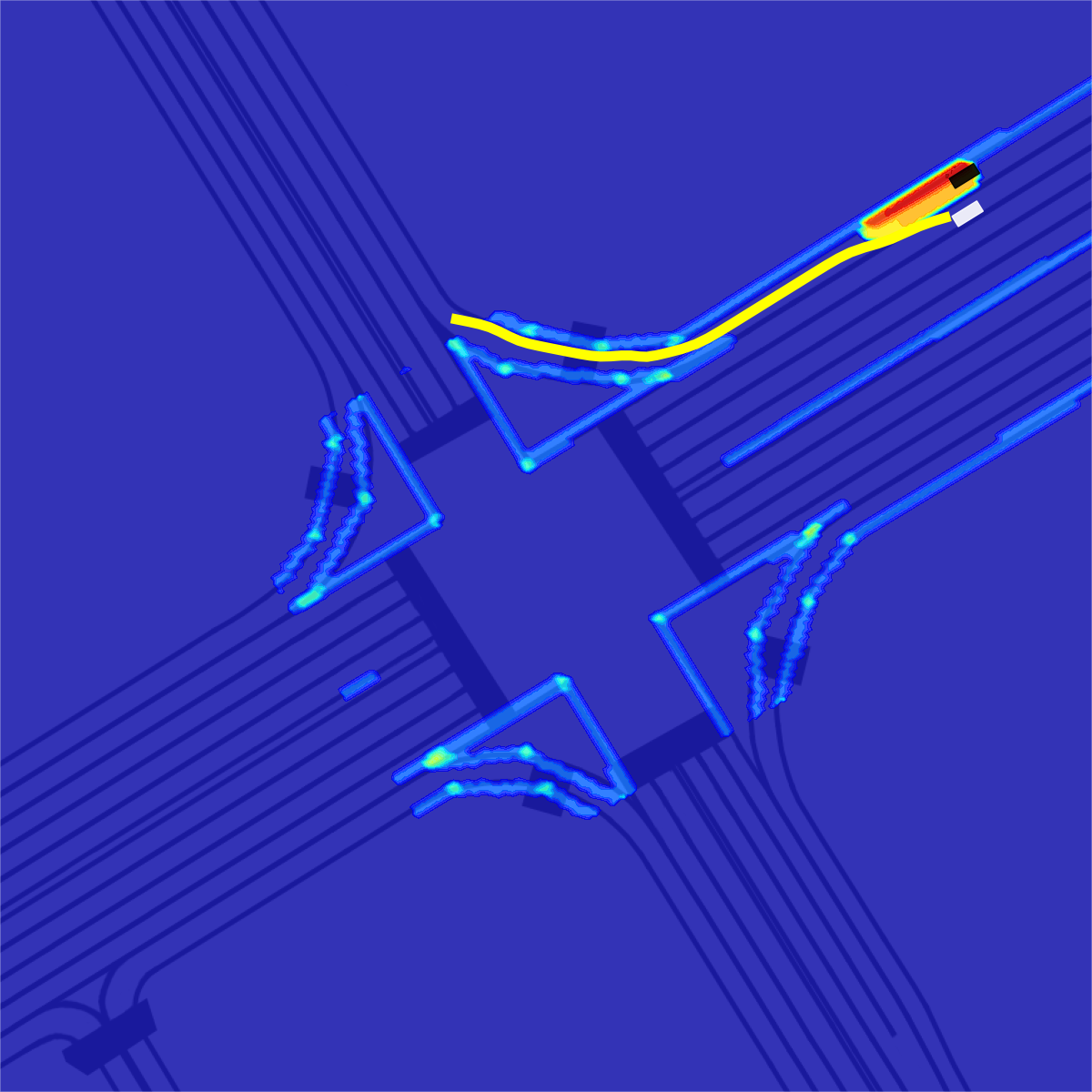}}
	 
	\end{minipage}
	\begin{minipage}{0.24\linewidth}
		\vspace{3pt}
		\centerline{\includegraphics[width=\textwidth]{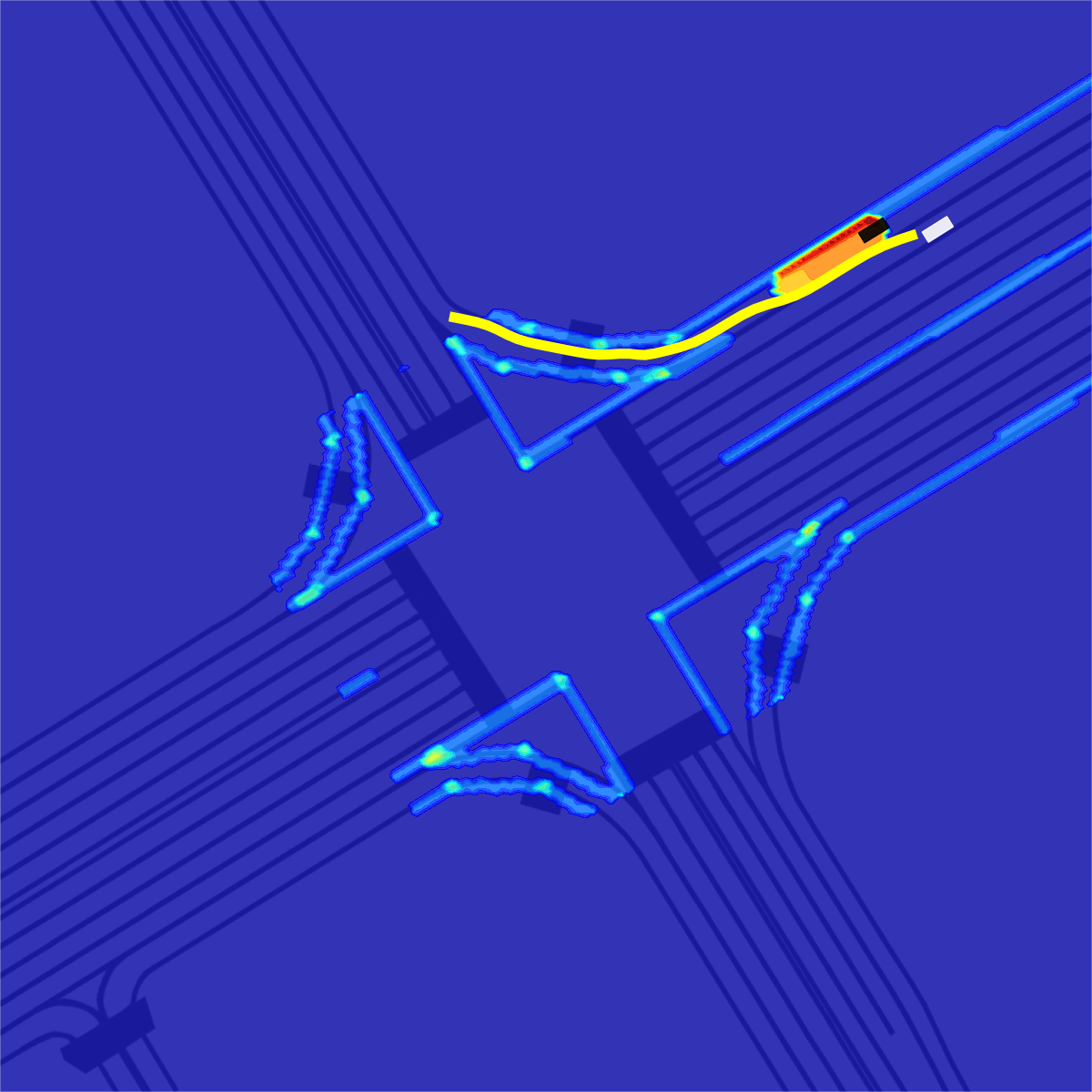}}
	 
	\end{minipage}
    \begin{minipage}{0.24\linewidth}
		\vspace{3pt}
		\centerline{\includegraphics[width=\textwidth]{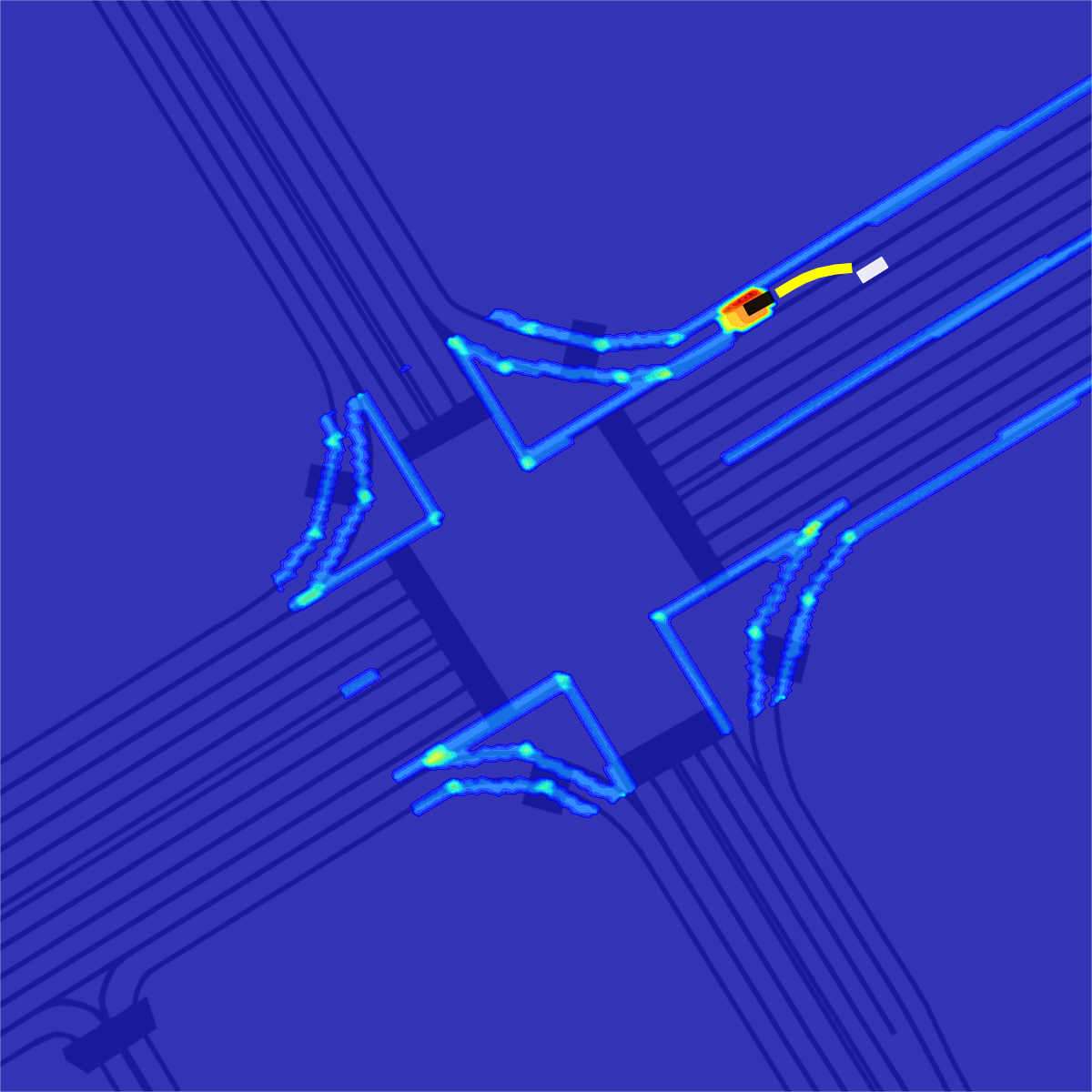}}
	\end{minipage}
 
    \raisebox{0\height}{\rotatebox{90}{\fontsize{9pt}{0pt}\selectfont Straight}}\hspace{0pt}
    \begin{minipage}{0.24\linewidth}
		\vspace{3pt}
		\centerline{\includegraphics[width=\textwidth]{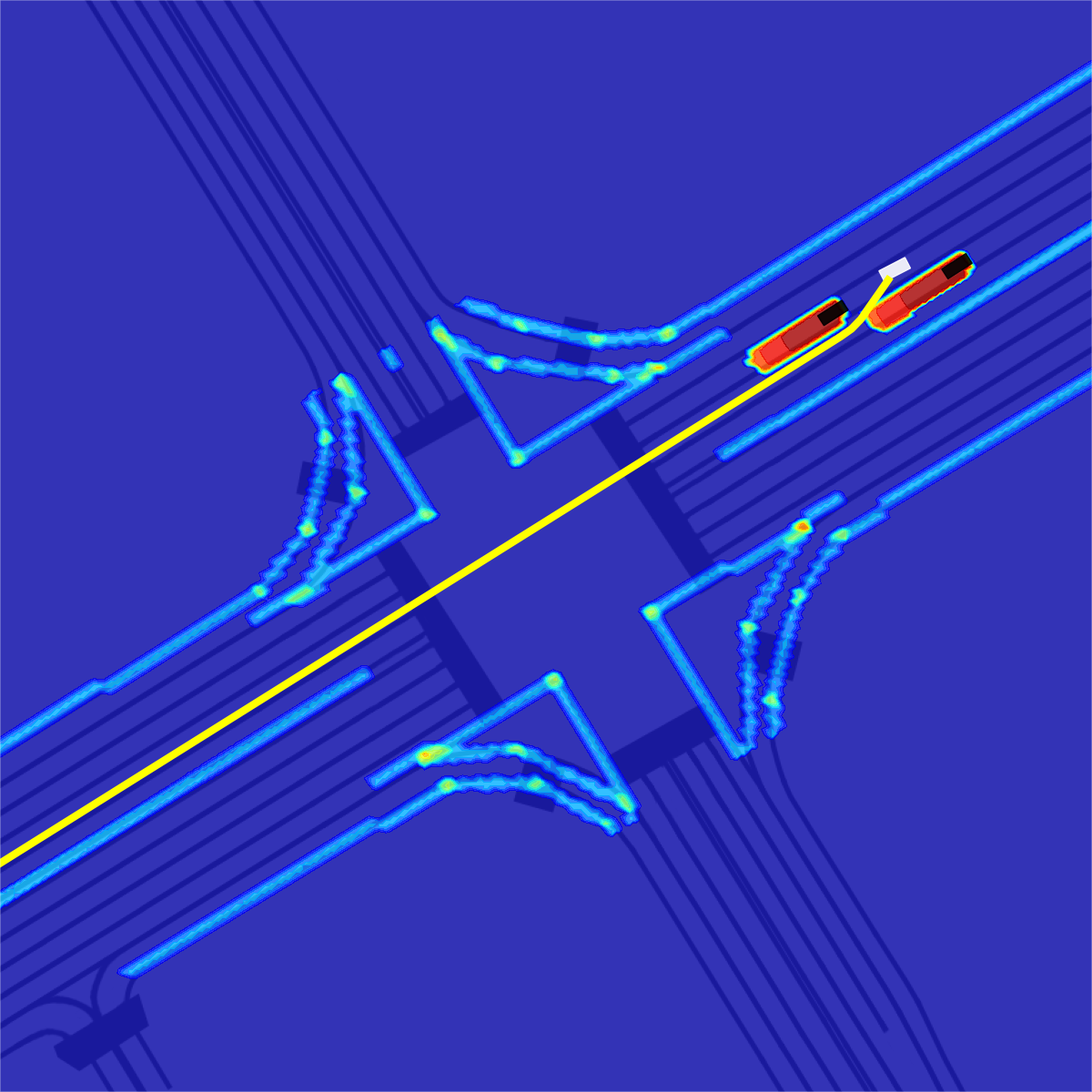}}
	 
	\end{minipage}
    \begin{minipage}{0.24\linewidth}
		\vspace{3pt}
		\centerline{\includegraphics[width=\textwidth]{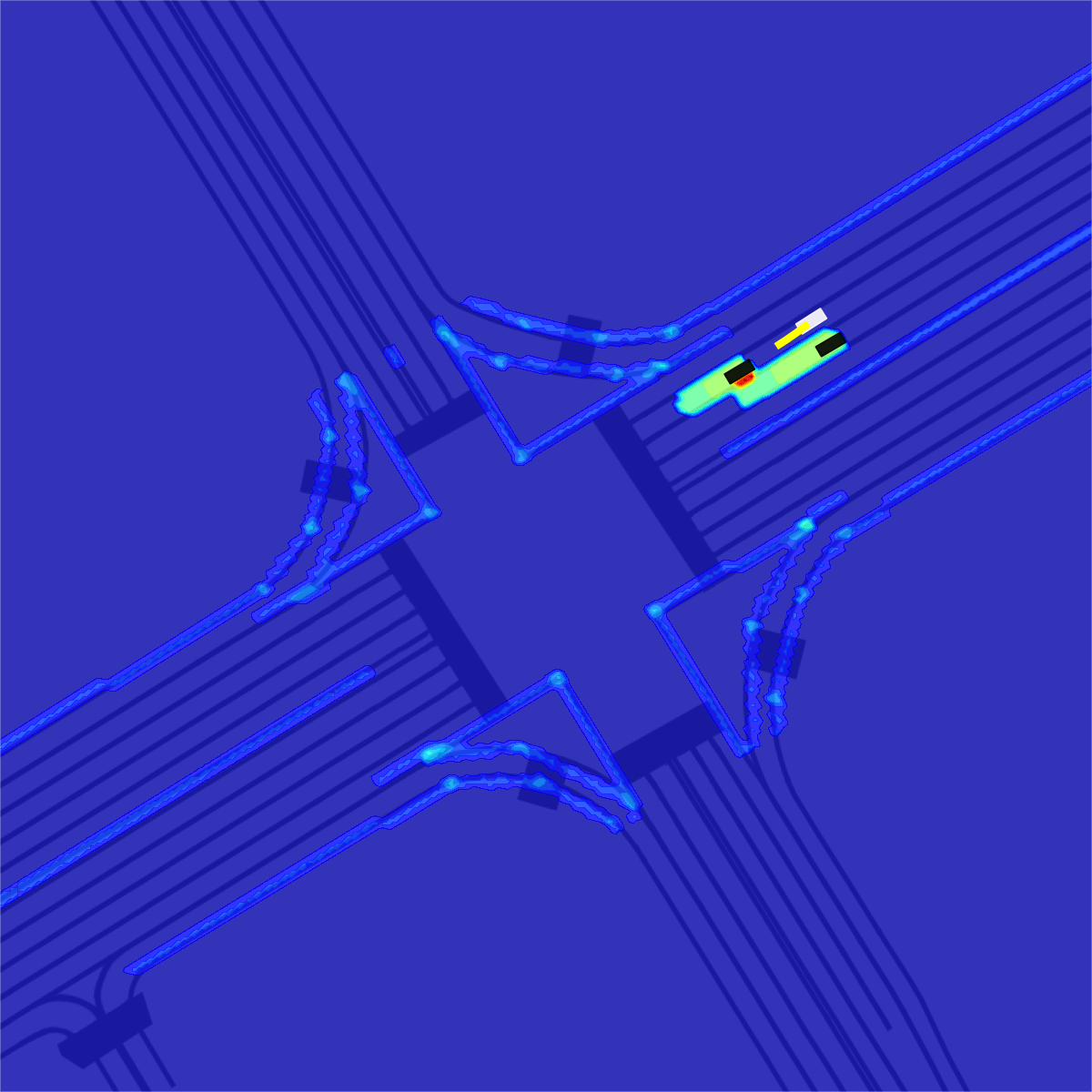}}
	 
	\end{minipage}
    \begin{minipage}{0.24\linewidth}
		\vspace{3pt}
		\centerline{\includegraphics[width=\textwidth]{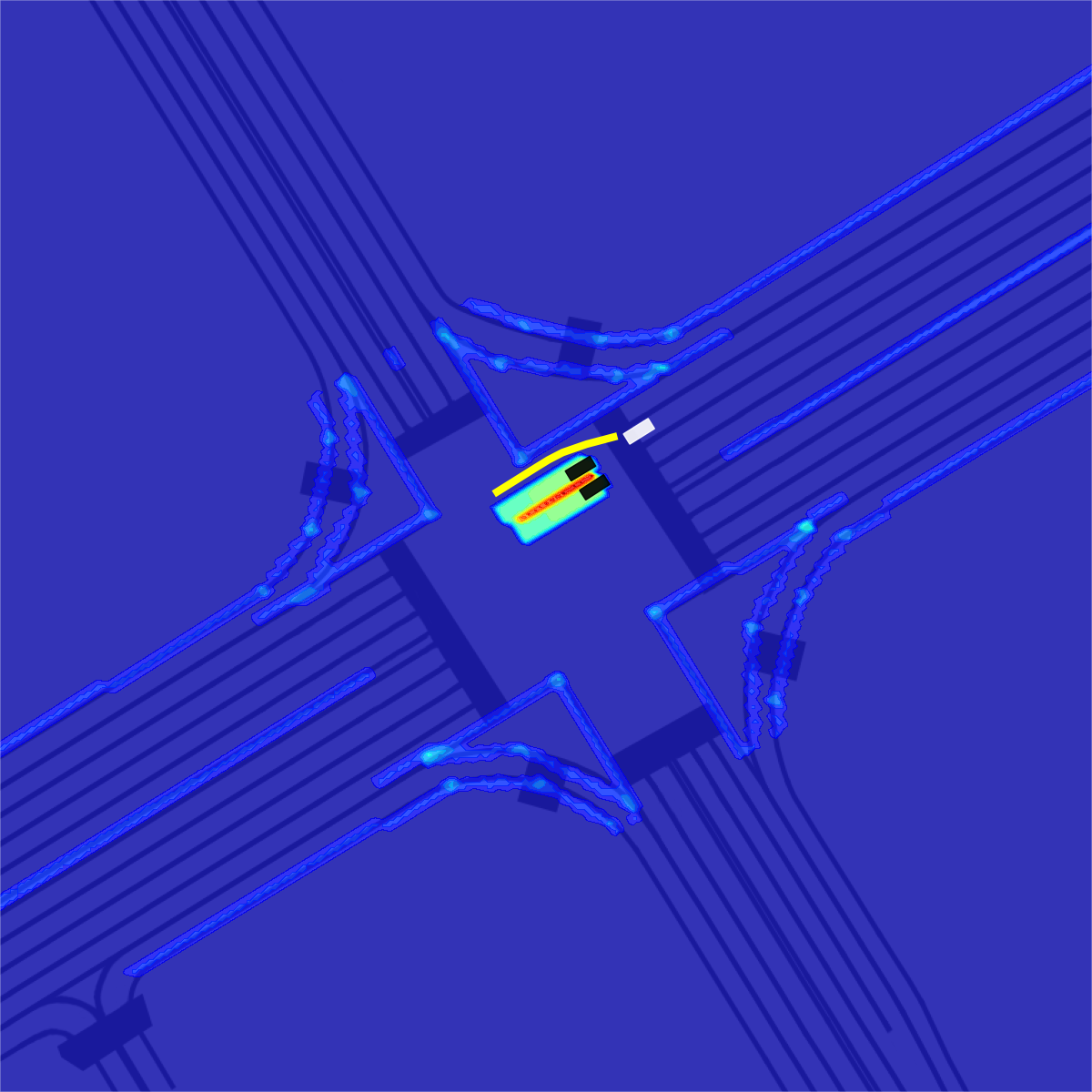}}
	 
	\end{minipage}
    \begin{minipage}{0.24\linewidth}
		\vspace{3pt}
		\centerline{\includegraphics[width=\textwidth]{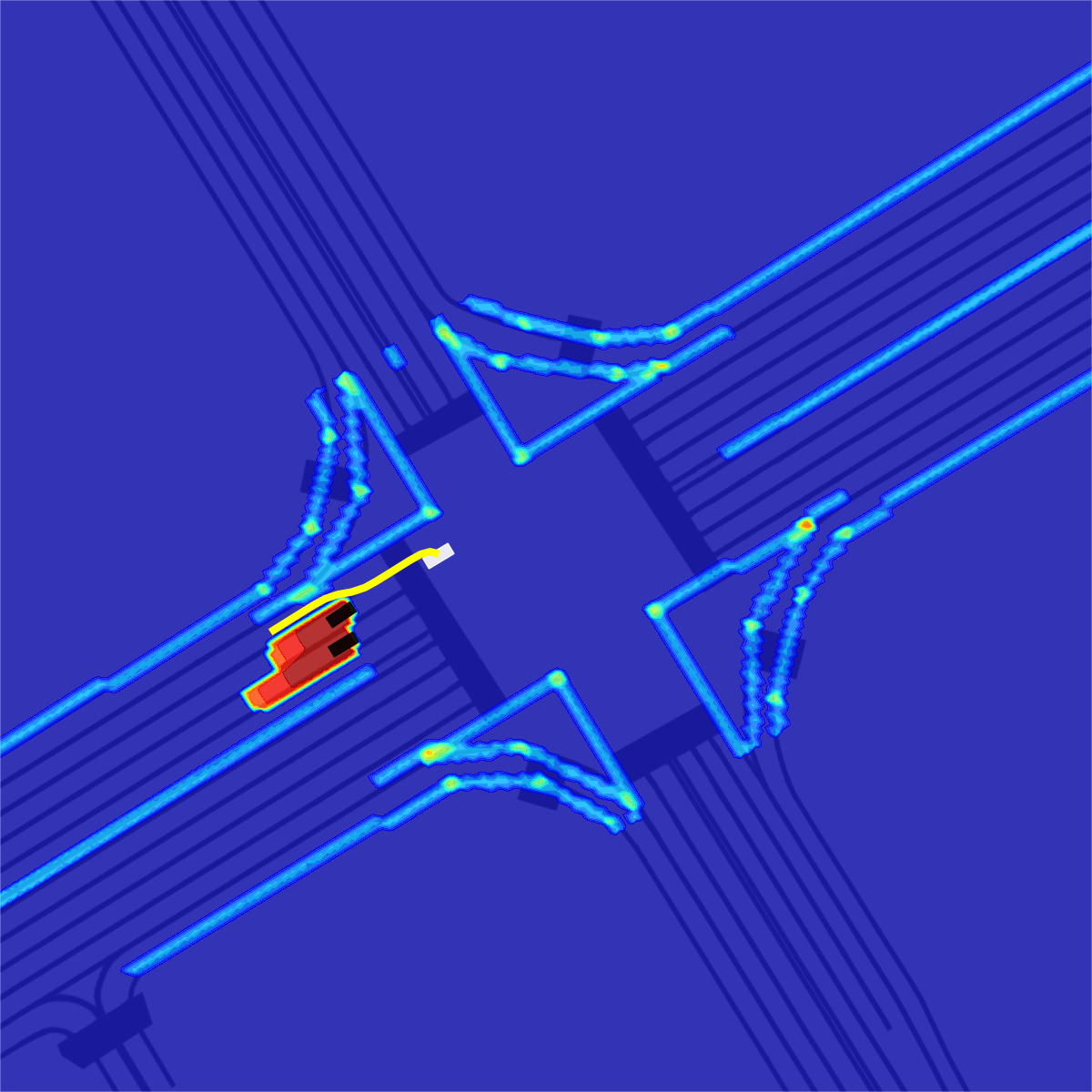}}
	 
	\end{minipage}

    \raisebox{0\height}{\rotatebox{90}{\fontsize{9pt}{0pt}\selectfont Left}}\hspace{2pt}
    \begin{minipage}{0.24\linewidth}
		\vspace{3pt}
		\centerline{\includegraphics[width=\textwidth]{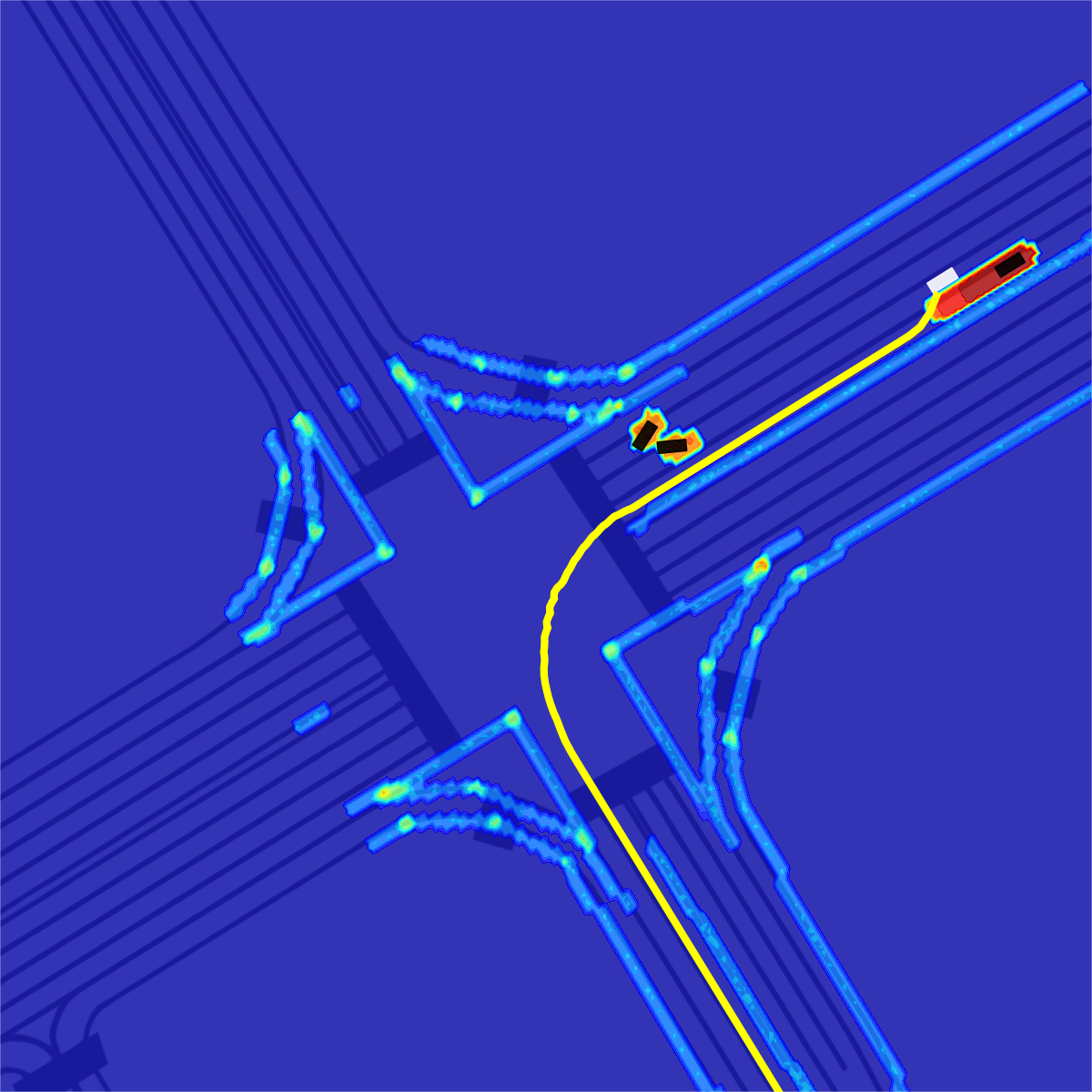}}
	 
	\end{minipage}
      \begin{minipage}{0.24\linewidth}
		\vspace{3pt}
		\centerline{\includegraphics[width=\textwidth]{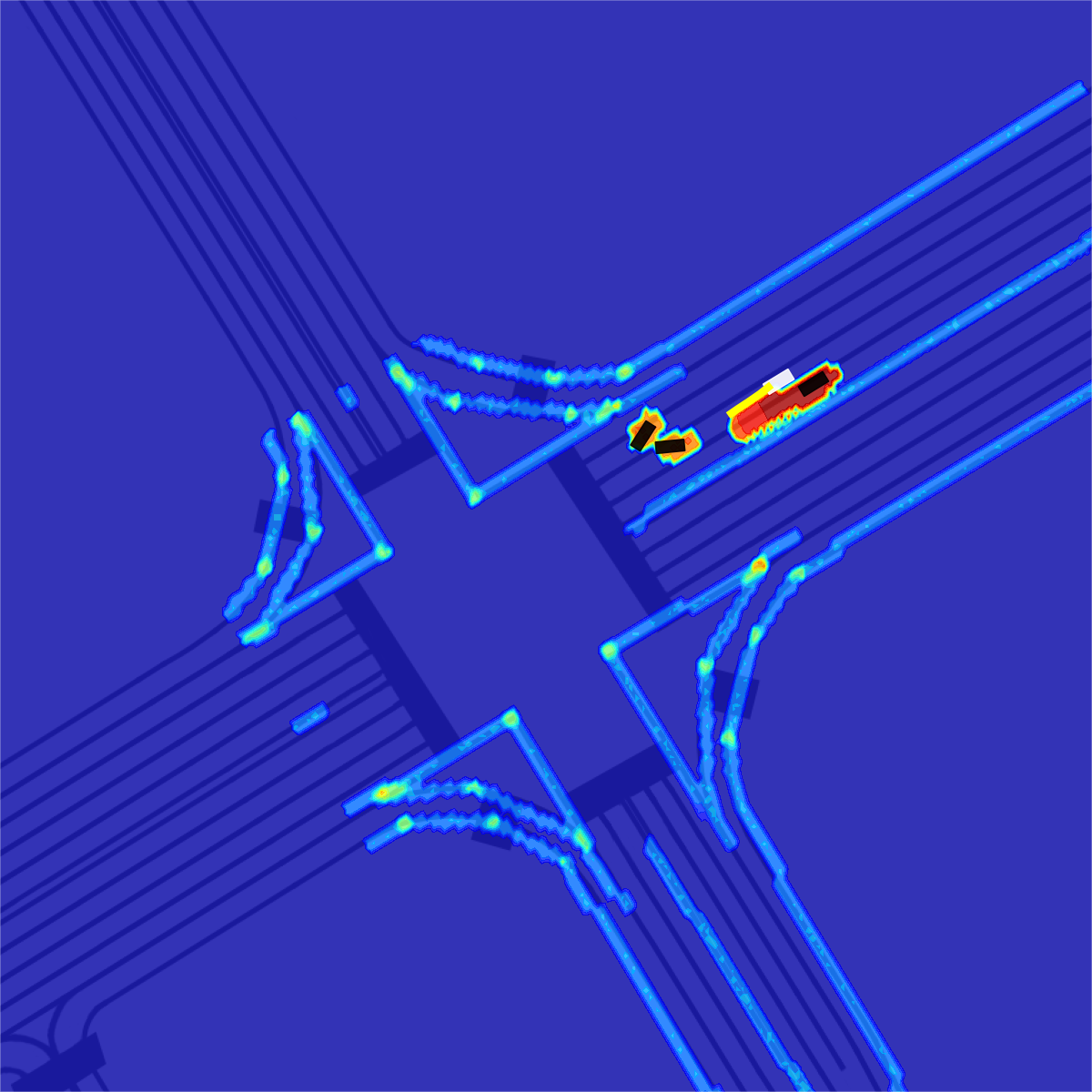}}
	 
	\end{minipage}
      \begin{minipage}{0.24\linewidth}
		\vspace{3pt}
		\centerline{\includegraphics[width=\textwidth]{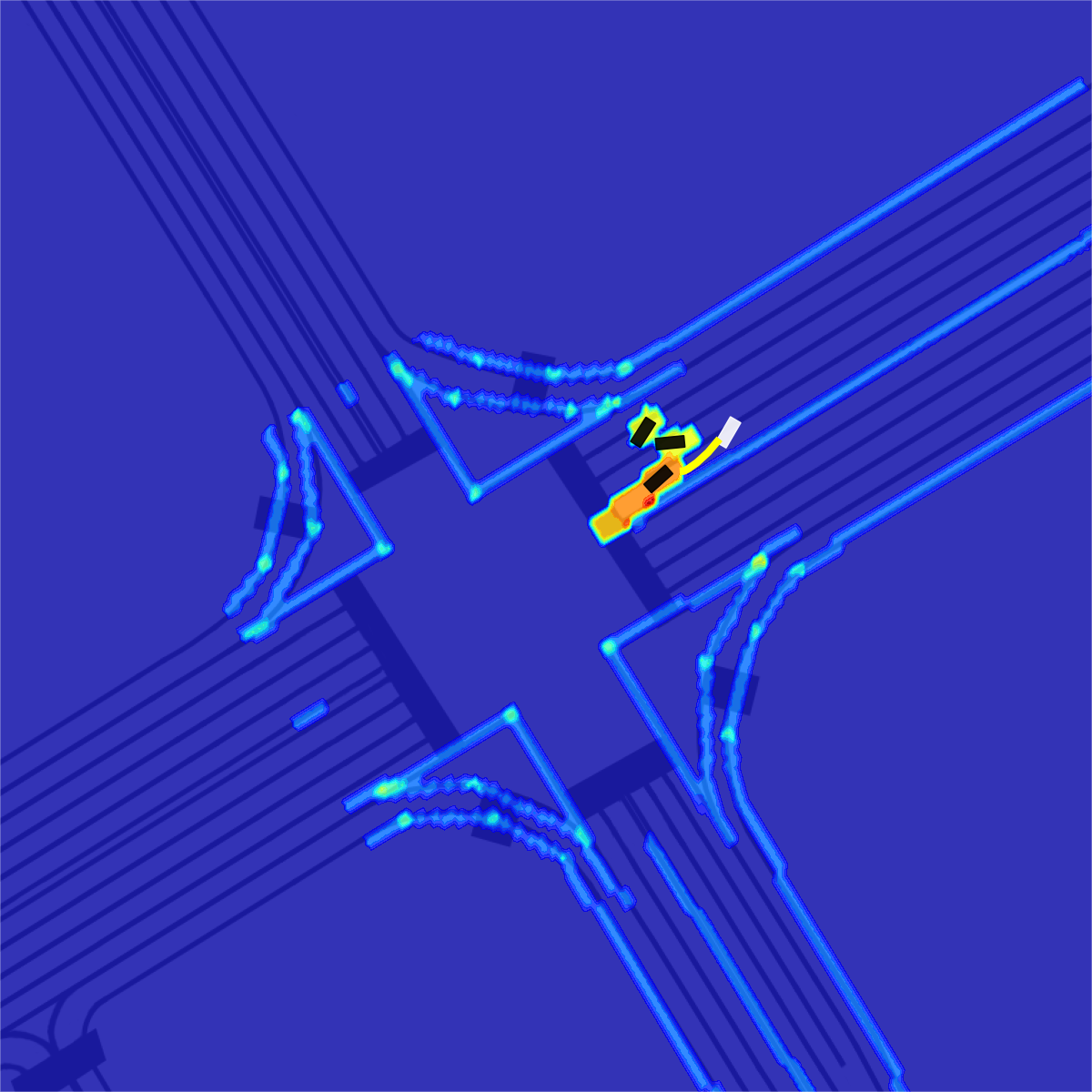}}
	 
	\end{minipage}
      \begin{minipage}{0.24\linewidth}
		\vspace{3pt}
		\centerline{\includegraphics[width=\textwidth]{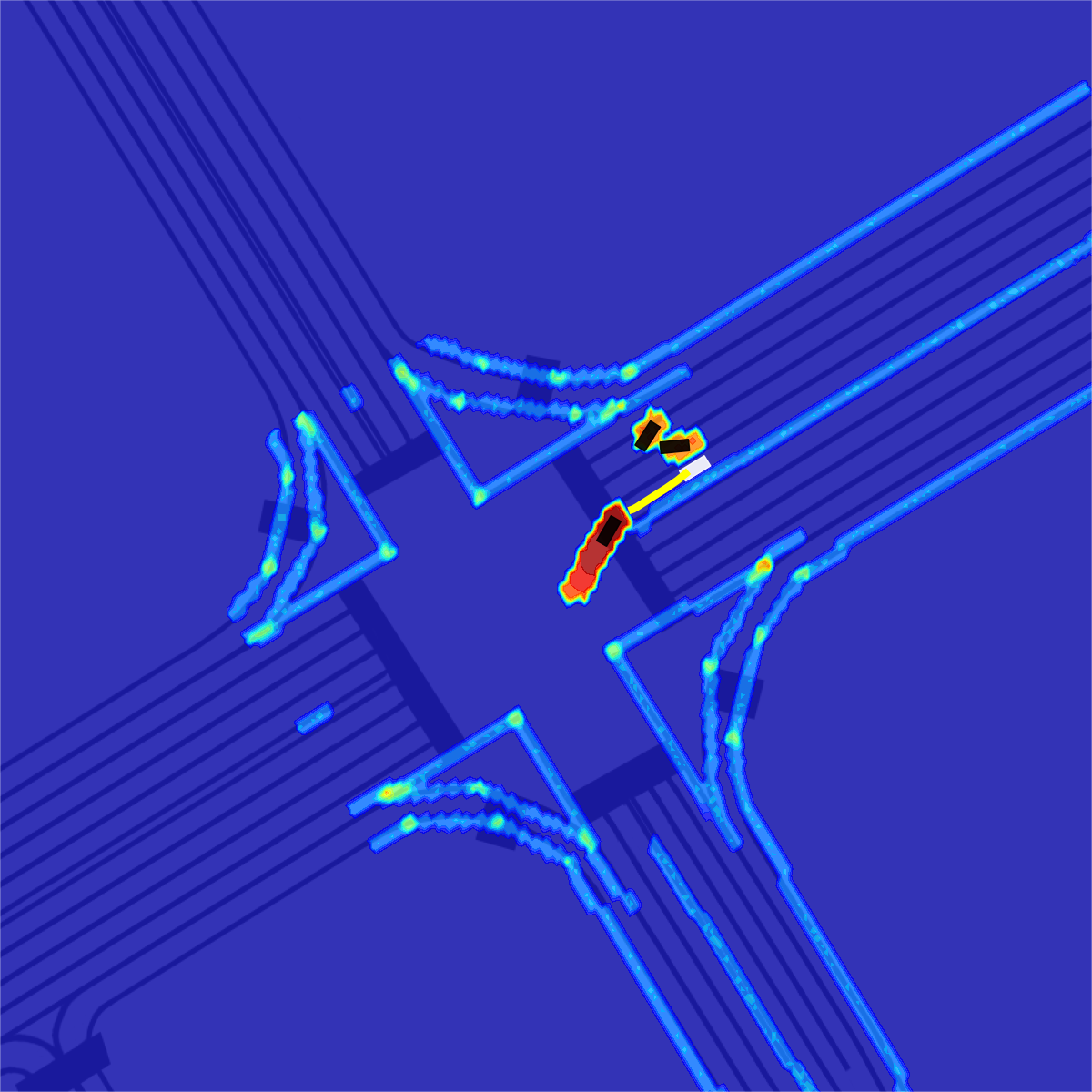}}
	 
	\end{minipage}

    \raisebox{0\height}{\rotatebox{90}{\fontsize{9pt}{0pt}\selectfont Left}}\hspace{2pt}
    \begin{minipage}{0.24\linewidth}
		\vspace{3pt}
		\centerline{\includegraphics[width=\textwidth]{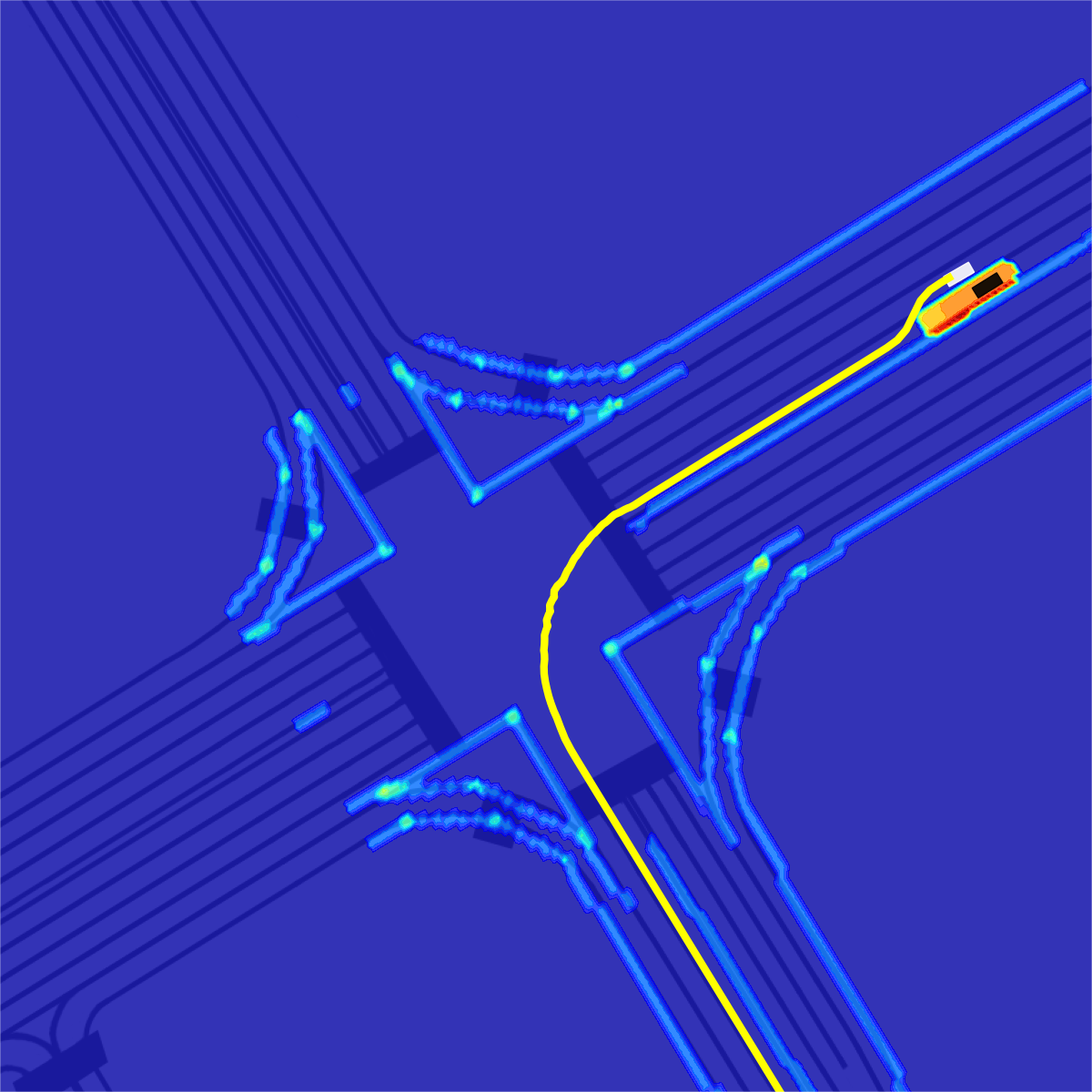}}
	 
		\centerline{step=1}
	\end{minipage}
     \begin{minipage}{0.24\linewidth}
		\vspace{3pt}
		\centerline{\includegraphics[width=\textwidth]{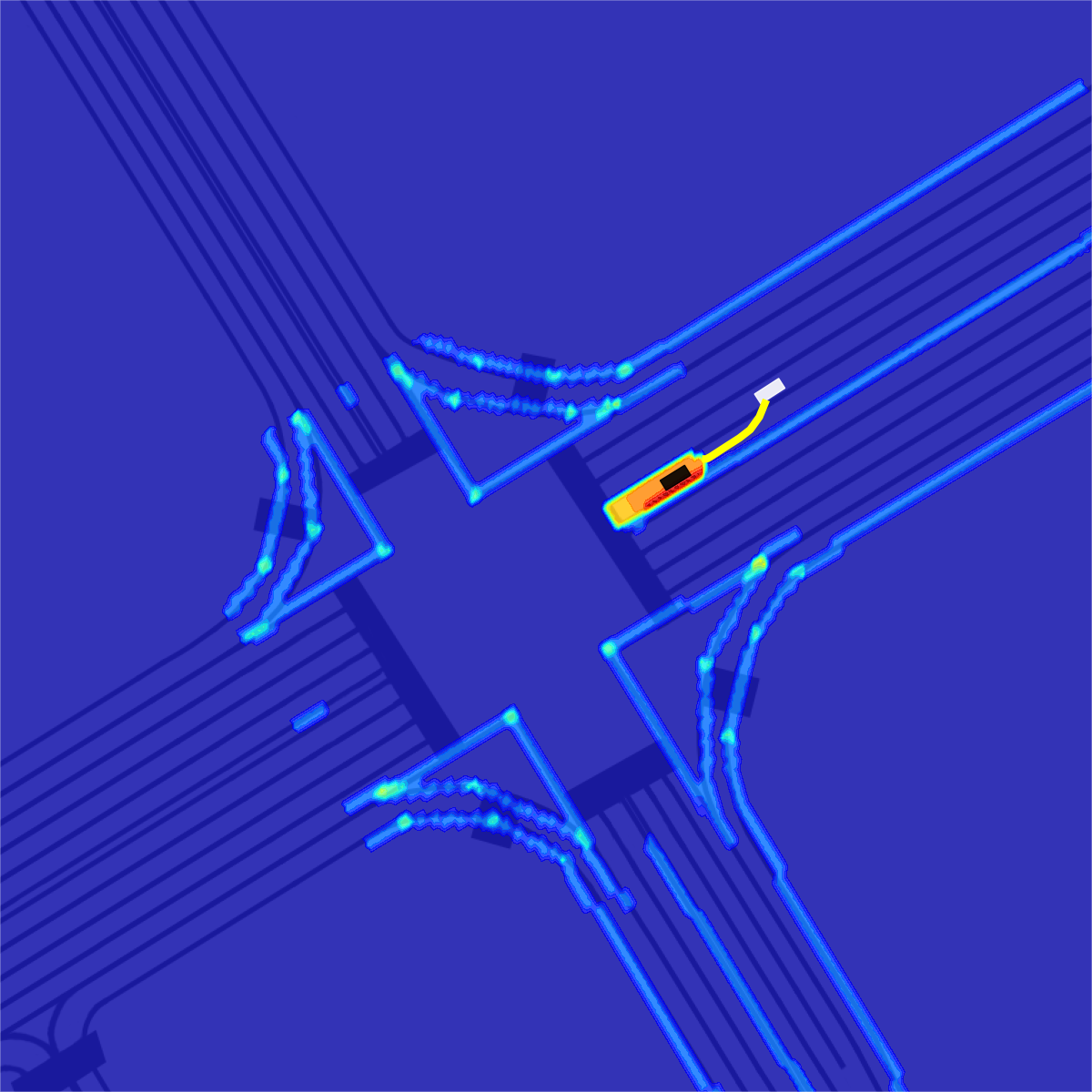}}
	 
		\centerline{step=2}
	\end{minipage}
     \begin{minipage}{0.24\linewidth}
		\vspace{3pt}
		\centerline{\includegraphics[width=\textwidth]{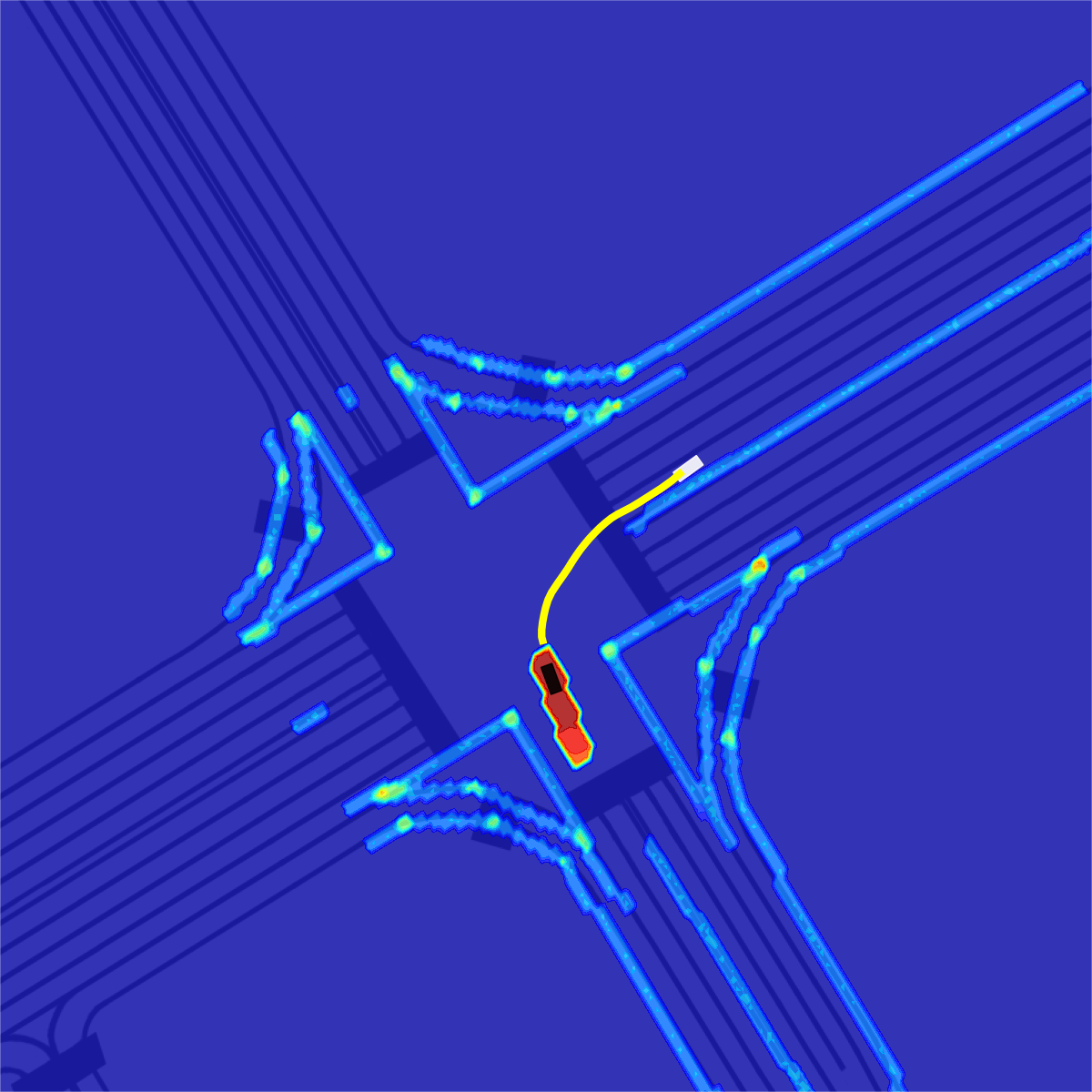}}
	 
		\centerline{step=3}
	\end{minipage}
     \begin{minipage}{0.24\linewidth}
		\vspace{3pt}
		\centerline{\includegraphics[width=\textwidth]{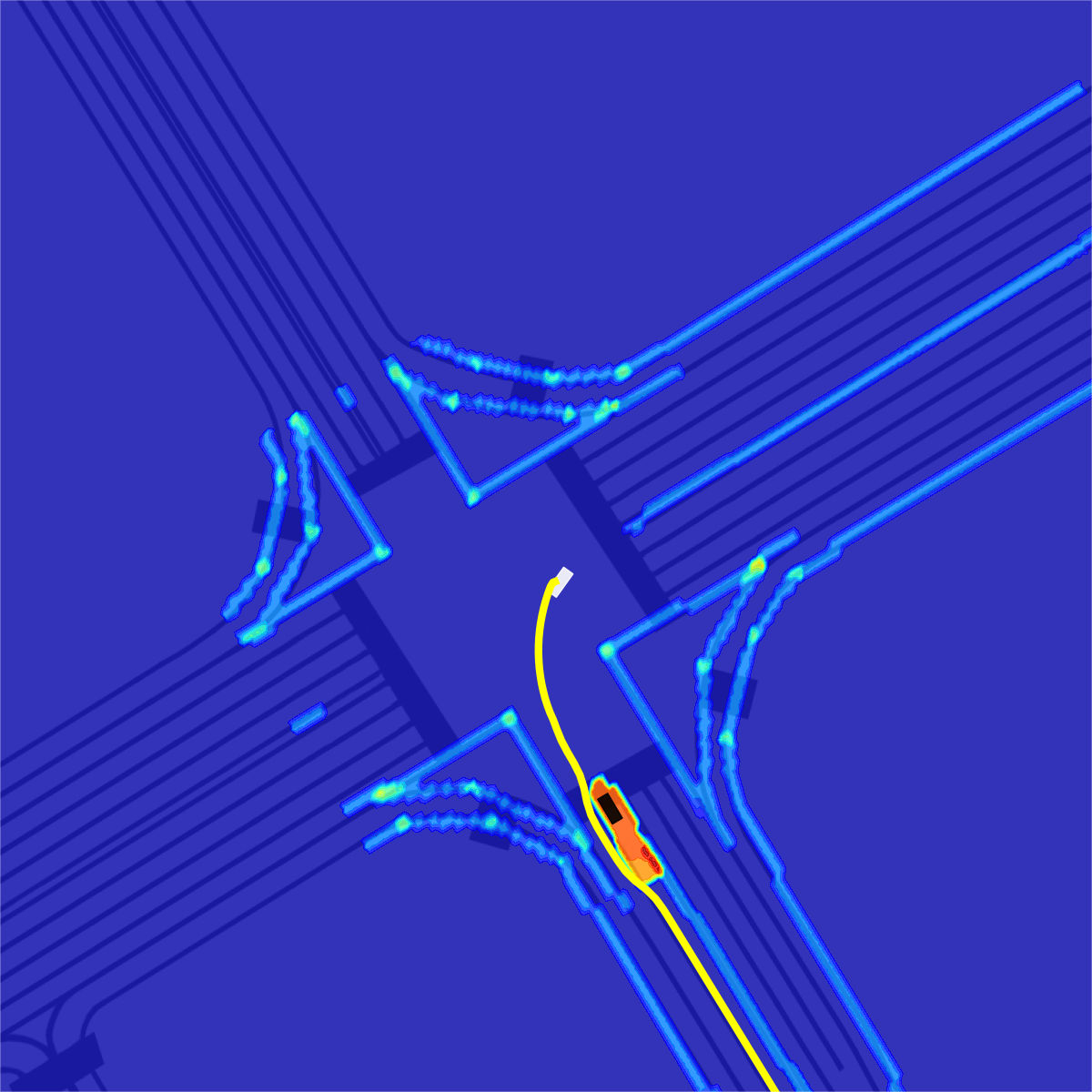}}
	 
		\centerline{step=4}
	\end{minipage}

    \centering
	\caption{Performance of the path planning algorithm based on 4D Risk Occupancy under various intersection-entry conditions.}
	\label{fig_experiment}
\end{figure*}

\section{Analysis of Experimental Results}

As shown in Fig. \ref{fig_experiment}, the 4D risk occupancy algorithm's feasibility and applicability were tested in a simulation experiment, where target ICVs and human-driven vehicles (HDVs) are represented by white and black rectangles, respectively. Road risk values are depicted using a gradient from dark blue to dark red, indicating increasing risk levels. 

\begin{table*}[t!]
\centering
\caption{Comparison of three scenarios}
\label{three}
\begin{tabular}{c|cccccc}
\toprule  \textbf{Scheme} & \textbf{Trajectory Planning} & \textbf{4D-Risk-Occ} & \textbf{Path Planning}  & \textbf{Maximum Safe Initial Speed}$\uparrow$  & \textbf{Average Acceleration} & \textbf{Behavior Pattern}\\ 
\midrule  \textbf{1}	& \checkmark & - & - &$5.42~\mathrm{m/s}$ & $-0.634~\mathrm{m/s^2}$ & Dangerous\\ 
\midrule  \textbf{2}	& \checkmark & \checkmark & - &$8.875~\mathrm{m/s}$  & $-0.536~\mathrm{m/s^{2}}$	&Normal \\
\midrule \textbf{3}	& \checkmark & \checkmark & \checkmark	&\textbf{$9.00~\mathrm{m/s}$} & \textbf{ $-0.507~\mathrm{m/s^{2}}$} & \textbf{Rule-Compliant}\\
\bottomrule
\end{tabular}
\end{table*}

The study further qualitatively analyzes the effectiveness of the path planning algorithm within a VRC collaborative framework through three scenarios for an ICV entering an intersection: a right turn, a straight movement, and a left turn, with yellow lines indicating the planned path. Four sequential frames of 4D Risk Occupancy visualization were extracted to show the algorithm's effectiveness under various traffic conditions. 
Furthermore, to quantitatively assess the advantages of 4D risk-occupancy-based path planning, a comparative analysis was conducted against a trajectory-planning algorithm using path guidance. This comparison aimed to evaluate the impact of using 4D risk occupancy and path-planning information compared with not using them.

\subsection{Qualitative Experiment}

\subsubsection{Experimental Analysis of ICV Right-Turn Behavior}

To validate the risk-occupancy-based path planning method for ICVs approaching intersections to make right turns, two test scenarios were designed, simulating different driving states.

The first row of Fig. \ref{fig_experiment} depicts the first right-turn scenario in four steps. The ICV, in the third lane from the right, prepares to turn right. An HDV overtakes from the far-right lane and enters the ramp first. From steps 1 to 3, the algorithm guides the ICV to change lanes twice, avoiding the HDV's risk area until step 4, when it adopts a cautious approach, guiding the ICV to drive straight until a safe distance from the HDV before turning right.
The second row shows the second scenario. The ICV, in the second lane from the right, needs to change lanes early. An HDV overtakes from the far-right lane and enters the ramp first. Initially, the algorithm suggests a direct lane change, but adjusts in steps 2 and 3, guiding the ICV to drive straight past the HDV's risk area before changing lanes. By step 4, as the HDV slows, the algorithm advises an immediate lane change to follow the HDV into the ramp.

\subsubsection{Experimental Analysis of ICV Straight-Through Behavior}

The third row in Fig. \ref{fig_experiment} shows an ICV traveling straight through an intersection. Before entering the intersection, the ICV is near HDVs traveling straight ahead and to the left rear, with relative positions changing over time. These four steps demonstrate the adaptability of the path planning algorithm in complex straight-ahead conditions.
In step 1, as the HDVs' risk occupancy regions begin to enclose the ICV, the algorithm recommends a lane change to the upper-left lane to overtake, utilizing road space to avoid restrictions. In step 2, with HDV risk-occupancy regions ahead and to the left, the ICV follows conservatively. In steps 3 and 4, after entering the intersection, the algorithm suggests overtaking the HDVs from the right and continuing straight in the right lane, which is less costly than bypassing from the left, as it reduces lane changes and conflict points.

\subsubsection{Experimental Analysis of ICV Left-Turn Behavior}

The fourth and fifth rows of images in Fig. \ref{fig_experiment} show that the ICV faces the risk posed by HDVs at its left rear when preparing to turn left into the intersection. The path planner recommends the following steps: First, the ICV changes to the upper-left lane to avoid HDVs' risk area and complete the overtaking maneuver; then it proceeds straight into the intersection and turns left. In the second and third steps, when HDVs overtake, the path planner advises the ICV to adopt a conservative approach: change lanes to the left, follow the HDVs while proceeding straight, and turn left at the appropriate time, avoiding risky overtaking. This strategy demonstrates that the path planner fully considers traffic rules and safety when making choices, thereby avoiding potential conflicts. The simulation results show that the path planning algorithm based on 4D risk occupancy can dynamically adjust strategies according to real-time traffic conditions, ensuring driving safety and compliance.

\subsection{Quantitative Experiment}

The experimental scenario involves an ICV making a left turn, with a roadblock ahead caused by a two-car collision, a vehicle traveling behind the ICV on the left, and a pedestrian jaywalking across the road, as shown in Table \ref{three}. This study employs a trajectory planner with target-path tracking capability \cite{2010} to analyze the impact of risk occupancy and trajectory planning algorithms on the safety of ICVs, with a maximum braking deceleration set at $-4.0~\mathrm{m/s^2}$.
Table \ref{three} and Fig. \ref{fig:enter-label} illustrate the differences among three scenarios:

\begin{figure}[h]

    \subfigure[Scheme 1]{\includegraphics[width=0.315\linewidth, height=0.30\linewidth]{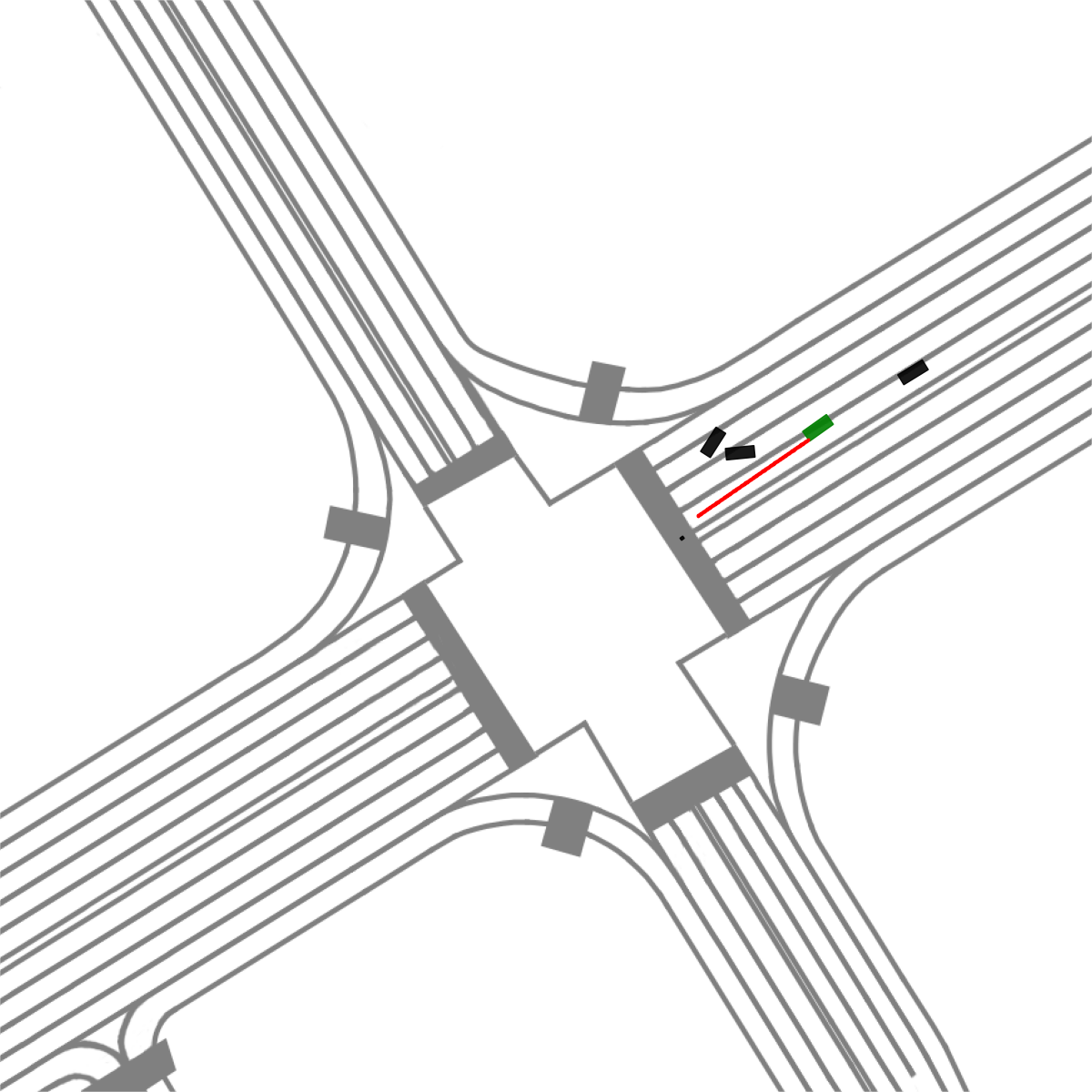}}
    \centering
    \subfigure[Scheme 2]{\includegraphics[width=0.315\linewidth,height=0.30\linewidth]{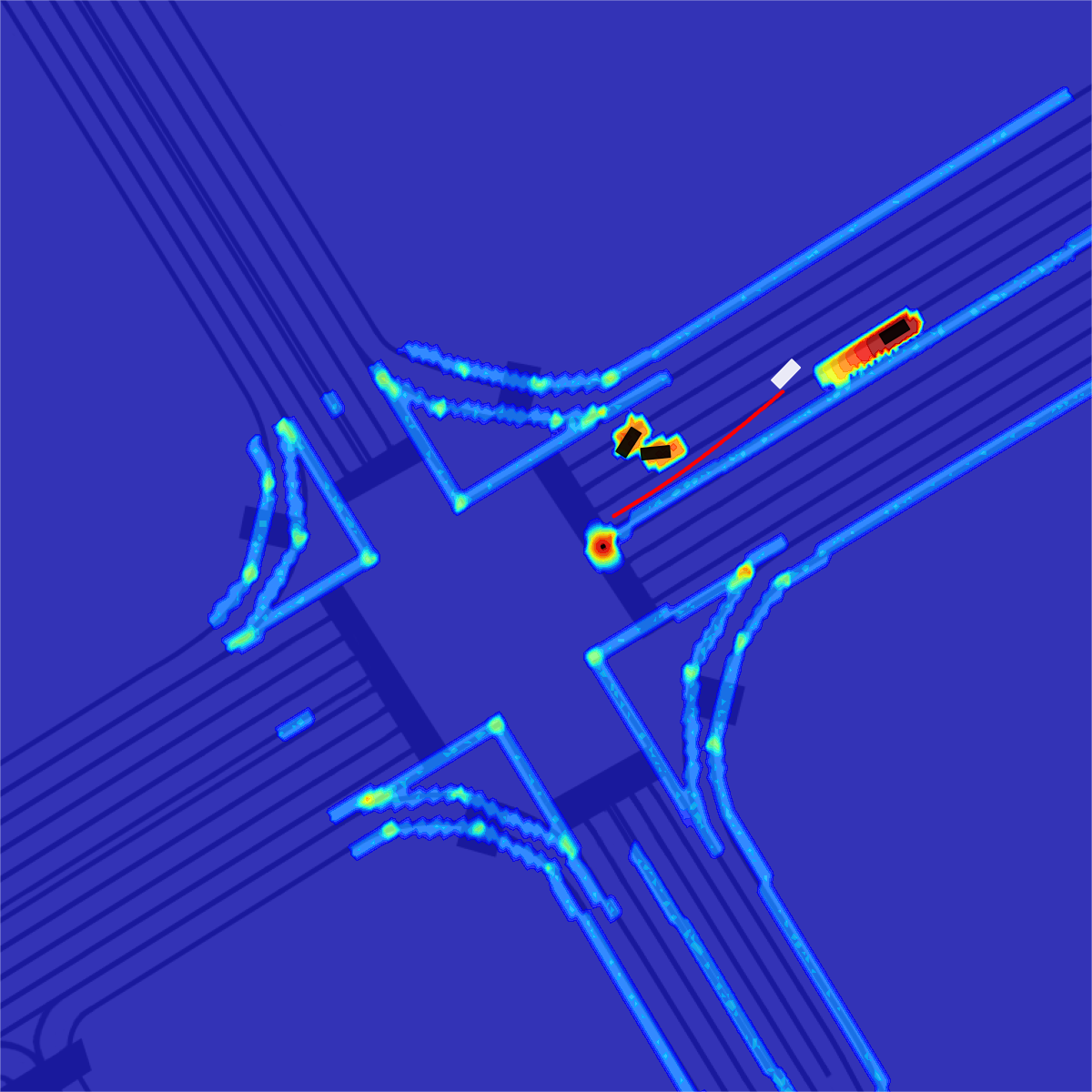}}
    \centering
    \subfigure[Scheme 3]{\includegraphics[width=0.315\linewidth,height=0.30\linewidth]{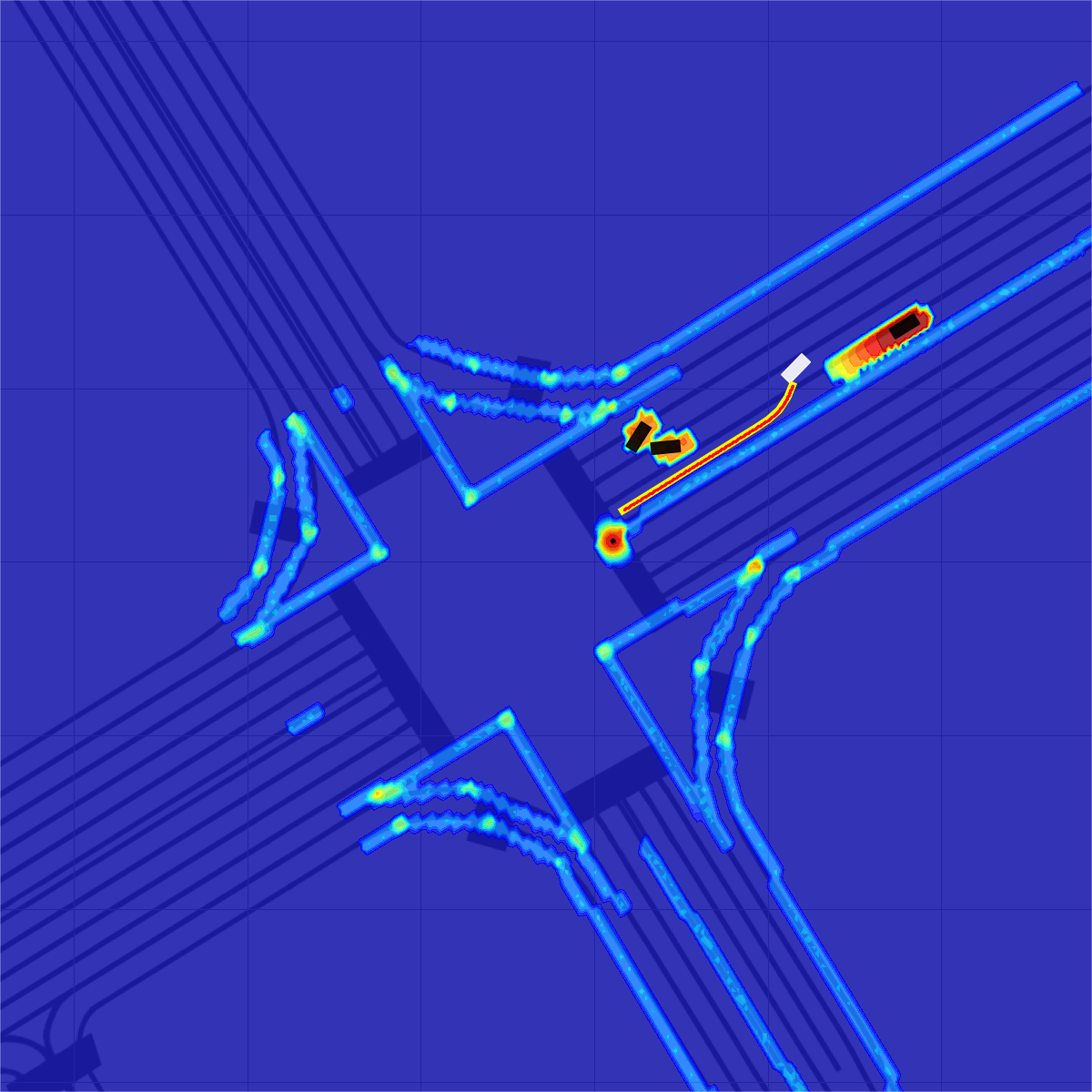}}
    \centering
    \caption{The trajectory-planning outcomes generated when the ICV detects the pedestrian are delineated by a sequence of red points for the three quantitative experimental schemes.}
    \label{fig:enter-label}
\end{figure}

\textbf{Scenario 1}: Only the trajectory planner is used, without risk occupancy or path-planning guidance. The ICV detects the pedestrian late, resulting in a short braking distance and a maximum safe initial speed of $5.42~\mathrm{m/s}$, with an average acceleration of $-0.634~\mathrm{m/s^2}$.

\textbf{Scenario 2}: The trajectory planner uses risk occupancy but not path-planning guidance. The ICV detects the pedestrian earlier and begins decelerating, with a maximum safe initial speed of $8.875~\mathrm{m/s}$. Using $8.0~\mathrm{m/s}$ as the initial braking speed, the average braking acceleration is $-0.536~\mathrm{m/s^2}$, but the lane-change completion time remains long.

\textbf{Scenario 3}: Both risk occupancy and path-planning functions are enabled. The maximum safe initial speed increases slightly to $9.0~\mathrm{m/s}$, and with an initial braking speed of $8.0~\mathrm{m/s}$, the average braking acceleration is $-0.507~\mathrm{m/s^2}$, which is more comfortable. The trajectory planning algorithm also significantly reduces lane-changing time and improves compliance with traffic rules.

The results show that trajectory planning guided by risk occupancy significantly enhances performance. While ensuring pedestrian safety, the maximum safe initial speed increases by $12.5\%$, and the magnitude of the average braking deceleration decreases by $5.41\%$ at an initial braking speed of $8.0~\mathrm{m/s}$, improving both safety and comfort.

\subsection{Comparative Experiment}

\begin{figure}[ht]
    \centering
    \includegraphics[width=\linewidth]{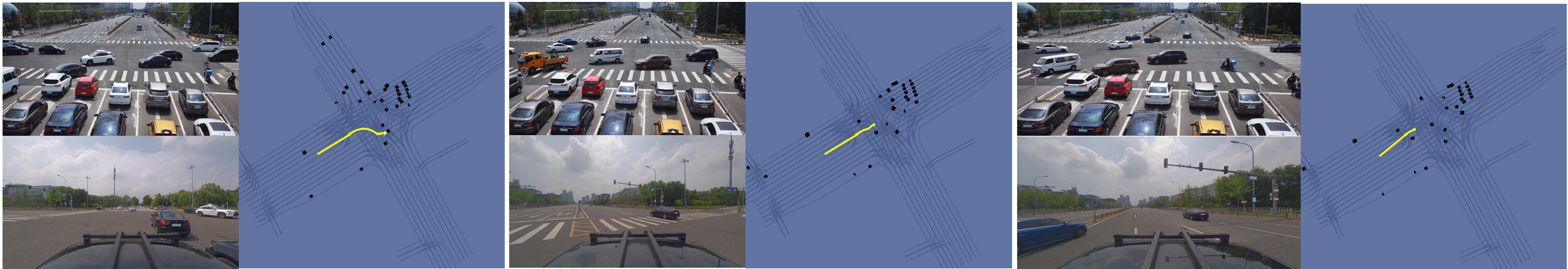}
    \caption{The visualization displays the roadside-perspective and vehicle-perspective images corresponding to three selected frames among the 130 frames, as well as the path planning results generated by our method.}
    \label{plan_ex}
\end{figure}

In addition to simulation experiments, a comparative study was conducted on the DAIR-V2X dataset. Specifically, a real-world intersection scenario was selected, 130 consecutive frames were extracted, and a left-turn path was planned for the ego ICV (Fig. \ref{plan_ex}), with the ICV's position as the start point and a fixed point in the post-turn lane as the end point in each frame. Other comparative algorithms were executed under identical conditions, with planning results statistically analyzed. Left-turn maneuvers involve threats from oncoming vehicles and interference from simultaneous left-turning vehicles. 

\begin{table}[ht]
    \centering
    \caption{Comparison of rule-based path planning methods}
    \label{plan_data}
    \setlength{\tabcolsep}{5pt}
    \begin{tabular}{c|ccccc}
    \toprule  \textbf{Method} & \textbf{A*} & \textbf{LPA*}	& \textbf{ARA*}  &  \textbf{Dijkstra} &  \textbf{Ours} \\ 
    \midrule  \textbf{Path Length}$\downarrow$	& 65.55 & 69.64 & 65.55 & 70.27 & \textbf{38.51}\\ 
    \midrule  \textbf{FPS}$\uparrow$	 & \textbf{166.7} & 32.3 & 100 & 124.6 & 1.4 / 274.3$\dagger$ \\ 
    \bottomrule
    \end{tabular}
    \begin{tablenotes}
    \footnotesize
    \item $\dagger$ $1.4~\mathrm{FPS}$ includes anchor network construction, risk evaluation, and path planning under anchor constraints, while $274.3~\mathrm{FPS}$ only covers the final anchor-connection phase.
    \end{tablenotes}
\end{table}

The proposed path planning algorithm is a heuristic method integrating prior information, risk occupancy maps, and anchor node constraints. Drawing on the cost-function logic of A*, multiple open-source A* variants and the classic rule-based Dijkstra algorithm were selected and adapted, with their generated path lengths compared to illustrate performance discrepancies. As shown in Table \ref{plan_data}, our method achieves the shortest path among the compared methods.
One limitation of the approach is its reliance on cloud-provided prior road information, pre-deployed anchor nodes, and post-processed upstream perception results.

\section{Conclusion}

This study introduces an innovative 4D Risk Occupancy algorithm to meet the driving-safety requirements of ICVs under a VRC collaborative architecture. The algorithm unifies directly observable perception results into a 4D Risk Occupancy Map. Additionally, we propose a method for using the Risk Occupancy Map, namely, a path-planning approach based on 4D risk occupancy and anchor node rules, to verify the effectiveness and applicability of 4D risk occupancy. 

Furthermore, the designed VRC collaborative application architecture ensures that the risk occupancy assessment algorithm can receive and process data from vehicles, roadside perception units, and edge clouds. Compared to perception paradigms such as object detection, local map construction, and occupancy grids, risk occupancy perception can offer a more comprehensive, continuous, unified, and high-level representation of information, allowing detailed differentiation of occupancy units and enriching perception representations under the VRC collaborative architecture.

\pagebreak
{\small
\bibliographystyle{unsrt}
\bibliography{main}
}
\pagebreak
\begin{IEEEbiography}
[{\includegraphics[width=1in,height=1.25in,clip,keepaspectratio]{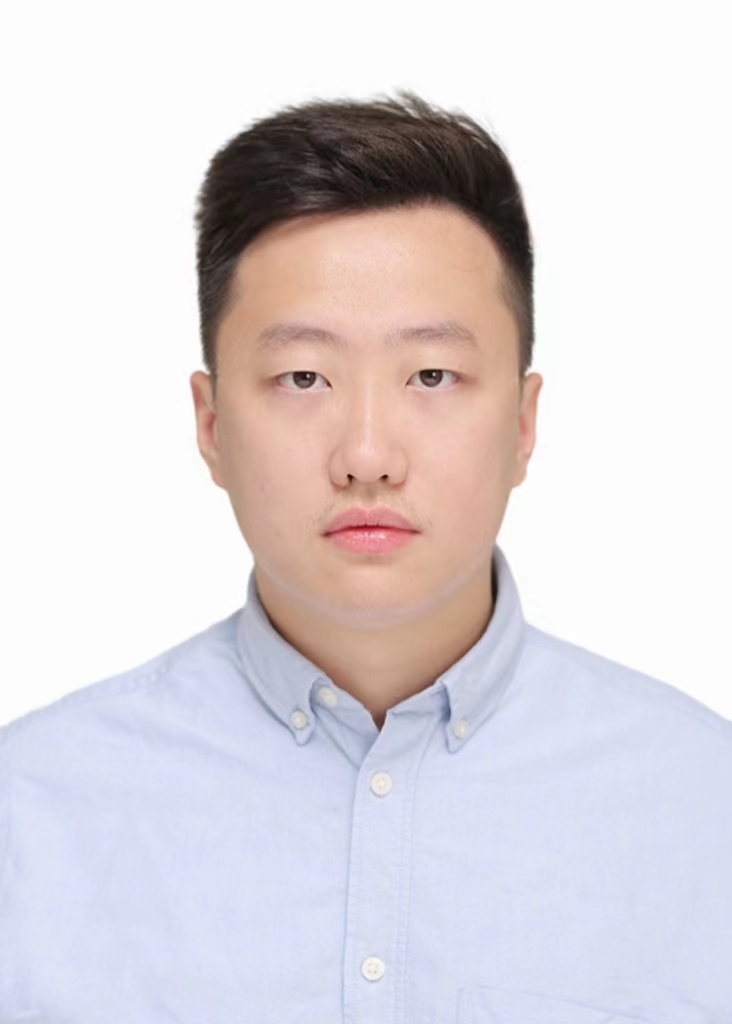}}]
{Jiaxing Chen} received an M.S. degree in Electrical and Computer Engineering from the University of Illinois, Chicago, USA, in 2021 and worked as an algorithm engineer at the National Innovation Center of Intelligent and Connected Vehicles, Beijing, China, from 2021 to 2022. He is currently pursuing a Ph.D. degree at the School of Vehicle and Mobility, Tsinghua University, Beijing, China. Chen’s research interests include computer vision and perception based on the Vehicle-Road-Cloud architecture.
\end{IEEEbiography}
\vspace{-10cm}

\begin{IEEEbiography}[{\includegraphics[width=1in,height=1.25in,clip,keepaspectratio]{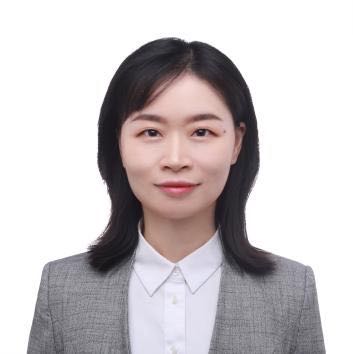}}]
{Wei Zhong} received the B.S. and Ph.D. degrees from Tsinghua University in 2010 and 2016, respectively. In 2013, she went to the University of Michigan (Ann Arbor) for a research visit at the Department of Mechanical Engineering. From 2019 to 2021, she conducted postdoctoral research at Tsinghua University. She is currently an Assistant Researcher at the State Key Laboratory of Intelligent Green Vehicle and Mobility at Tsinghua University. Her research interests mainly focus on Intelligent Connected Vehicles, including system architecture, multi-objective optimization, and cloud control platforms.
\end{IEEEbiography}
\vspace{-10cm}

\begin{IEEEbiography}
[{\includegraphics[width=1in,height=1.25in,clip,keepaspectratio]{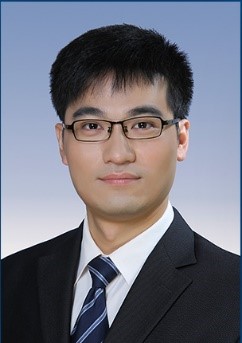}}]
{Bolin Gao} received the B.E. and M.E. degrees in Vehicle Engineering from Jilin University, Changchun, China, in 2007 and 2009, respectively, and the Ph.D. degree in Vehicle Engineering from Tongji University, Shanghai, China, in 2013. He is currently an associate research professor at the School of Vehicle and Mobility, Tsinghua University. His research interests include the theoretical research and engineering application of the dynamic design and control of Intelligent and Connected Vehicles, especially collaborative perception and tracking methods in cloud control systems, intelligent predictive cruise control systems on commercial trucks with cloud control mode, as well as testing and evaluation of intelligent vehicle driving systems.
\end{IEEEbiography}
\vspace{-10cm}

\begin{IEEEbiography}[{\includegraphics[width=1in,height=1.25in,clip,keepaspectratio]{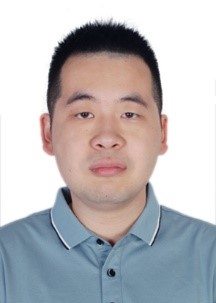}}]
{Yifei Liu} received his bachelor's degree from Huazhong University of Science and Technology, China, in 2020 and his master's degree from the University of Bristol, United Kingdom, in 2023. He is currently working as an engineer in the Intelligent Connected Vehicles Research Group at the School of Vehicle and Mobility, Tsinghua University. His research focuses on road section perception and fusion under the Vehicle-Road-Cloud integrated architecture.
\end{IEEEbiography}
\vspace{-10cm}

\begin{IEEEbiography}[{\includegraphics[width=1in,height=1.25in,clip,keepaspectratio]{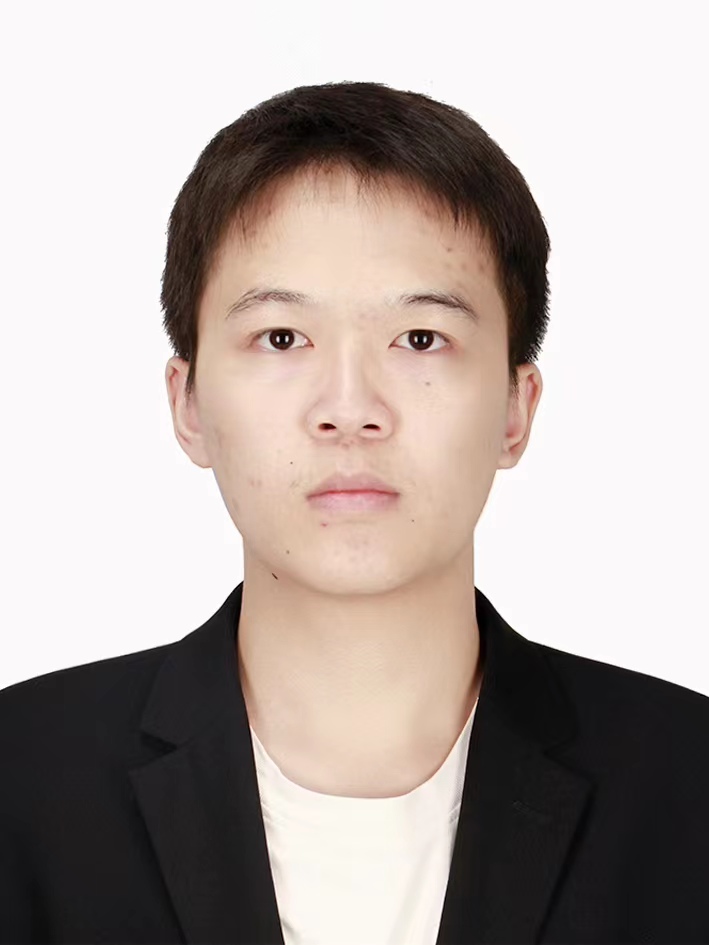}}]
{Hengduo Zou} is pursuing a B.E. degree in Mechanical Engineering at the School of Vehicle and Mobility, Tsinghua University. His research interests include autonomous driving, computer vision, and Vehicle-Road-Cloud integration systems.
\end{IEEEbiography}

\vspace{-10cm}

\newpage

\begin{IEEEbiography}[{\includegraphics[width=1in,height=1.25in,clip,keepaspectratio]{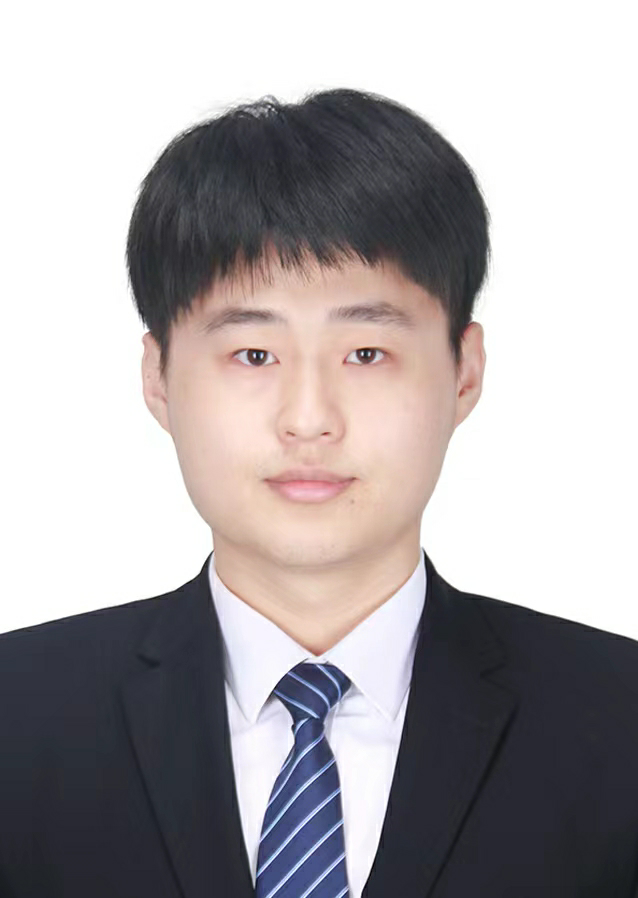}}]
{Jiaxi Liu} received the B.E. and M.E. degrees in Mechanical Engineering from the School of Vehicle and Mobility, Tsinghua University. He is now pursuing a Ph.D. degree in civil engineering at the University of Wisconsin-Madison. His research interests include collaborative perception, real-time perception, Vehicle-Road-Cloud integration systems, and cyber-physical system architecture design.
\end{IEEEbiography}
\vspace{-5cm}

\begin{IEEEbiography}[{\includegraphics[width=1in,height=1.25in,clip,keepaspectratio]{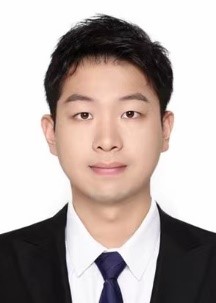}}]
{Yanbo Lu} received the B.S. degree in Mechanical Engineering from Nanjing University of Aeronautics and Astronautics, China, in 2017, the M.S. degree in Mechanical Engineering from Embry-Riddle Aeronautical University, USA, in 2018, and the Ph.D. degree in Mechanical Engineering from Southeast University, China, in 2023. He is currently a postdoctoral research fellow with the School of Vehicle and Mobility, Tsinghua University. His research interests include vehicle dynamics and control, Vehicle-Cloud control systems and fault-tolerant control.
\end{IEEEbiography}
\vspace{-5cm}

\begin{IEEEbiography}[{\includegraphics[width=1in,height=1.25in,clip,keepaspectratio]{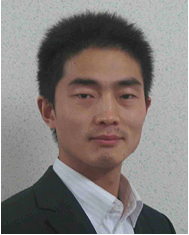}}]
{Jin Huang} received the B.E. and Ph.D. degrees from
 the College of Mechanical and Vehicle Engineering,
 Hunan University, Changsha, China, in 2006 and 2012, respectively. From 2009 to 2011, he was also a joint Ph.D. student with the George W. Woodruff School of
 Mechanical Engineering, Georgia Institute of Technology, Atlanta, GA, USA. He started his career as a postdoctoral researcher and an assistant research professor at Tsinghua University, Beijing, China, in 2013 and 2016,
 respectively. His research interests include artificial
 intelligence in intelligent transportation systems, dynamics control, and fuzzy engineering.
\end{IEEEbiography}
\vspace{-5cm}

\begin{IEEEbiography}[{\includegraphics[width=1in,height=1.25in,clip,keepaspectratio]{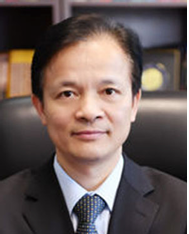}}]
{Zhihua Zhong} received a Ph.D. degree in engineering from Linköping University, Linköping, Sweden,
in 1988. He is currently a Professor at the School
of Vehicle and Mobility, Tsinghua University, Beijing, China. He was an elected member of the Chinese
 Academy of Engineering in 2005. His research interests include automotive collision safety technology, automotive body stamping and forming technologies,
 modularity and automotive lightweighting technologies, and
 vehicle dynamics.
\end{IEEEbiography}
\vspace{-5cm}

\end{document}